\documentclass[acmtog, nonacm]{acmart}

\usepackage{booktabs} % For formal tables
\usepackage[ruled]{algorithm2e} % For algorithms

\SetAlFnt{\small}
\SetAlCapFnt{\small}
\SetAlCapNameFnt{\small}
\SetAlCapHSkip{0pt}

\usepackage{xcolor} 
\usepackage{colortbl}
\usepackage{float}
\definecolor{best1}{RGB}{222,242,212}
\definecolor{best2}{RGB}{255,250,212}

\usepackage{bbding}

\usepackage{color,hyperref}
\usepackage{markdown}

\begin{document}
\author{Jiakai Zhang}
\orcid{0000-0001-9477-3159}
\affiliation{%
	\institution{ShanghaiTech University}
	\city{Shanghai}
	\country{China}}
\affiliation{%
	\institution{Stereye Intelligent Technology Co.,Ltd.}
	%\city{Shanghai}
	\country{China}}
\email{zhangjk@shanghaitech.edu.cn}

\author{Liao Wang}
\affiliation{%
	\institution{ShanghaiTech University}
	\city{Shanghai}
	\country{China}
}
\email{wangla@shanghaitech.edu.cn}

\author{Xinhang Liu}
\affiliation{%
	\institution{ShanghaiTech University}
	\city{Shanghai}
	\country{China}
}
\email{liuxh2@shanghaitech.edu.cn}

\author{Fuqiang Zhao}
\affiliation{%
	\institution{ShanghaiTech University}
	\city{Shanghai}
	\country{China}
}
\email{zhaofq@shanghaitech.edu.cn}

\author{Minzhang Li}
\affiliation{%
	\institution{ShanghaiTech University}
	\city{Shanghai}
	\country{China}
}
\email{limzh@shanghaitech.edu.cn}

\author{Haizhao Dai}
\affiliation{%
	\institution{ShanghaiTech University}
	\city{Shanghai}
	\country{China}
}
\email{daihzh@shanghaitech.edu.cn}

\author{Boyuan Zhang}
\affiliation{%
	\institution{ShanghaiTech University}
	\city{Shanghai}
	\country{China}
}
\email{zhangby@shanghaitech.edu.cn}

\author{Wei Yang}
\affiliation{%
	\institution{Huazhong University of Science and Technology}
	\city{Wuhan}
	\country{China}
}
\email{weiyangcs@hust.edu.cn}

\author{Lan Xu}
\affiliation{%
	\institution{ShanghaiTech University}
	\city{Shanghai}
	\country{China}
}
\email{xulan1@shanghaitech.edu.cn}

\author{Jingyi Yu}
\affiliation{%
	\institution{ShanghaiTech University}
	\city{Shanghai}
	\country{China}
}
\email{yujingyi@shanghaitech.edu.cn}
\authornote{The corresponding author is Jingyi Yu (yujingyi@shanghaitech.edu.cn). }

	% Title portion
	\title{NeuVV: Neural Volumetric Videos with Immersive Rendering and Editing}

	\begin{abstract}
		Some of the most exciting experiences that Metaverse promises to offer, for instance, live interactions with virtual characters in virtual environments, require real-time photo-realistic rendering. 3D reconstruction approaches to rendering, active or passive, still require extensive cleanup work to fix the meshes or point clouds. In this paper, we present a neural volumography technique called neural volumetric video or NeuVV to support immersive, interactive, and spatial-temporal rendering of volumetric video contents with photo-realism and in real-time. The core of NeuVV is to efficiently encode a dynamic neural radiance field (NeRF)~\cite{mildenhall2020nerf} into renderable and editable primitives. We introduce two types of factorization schemes: a hyper-spherical harmonics (HH) decomposition for modeling smooth color variations over space and time and a learnable basis representation for modeling abrupt density and color changes caused by motion. NeuVV factorization can be integrated into a Video Octree (VOctree) analogous to PlenOctree~\cite{yu2021plenoctrees} to significantly accelerate training while reducing memory overhead. Real-time NeuVV rendering further enables a class of immersive content editing tools. Specifically, NeuVV treats each VOctree as a primitive and implements volume-based depth ordering and alpha blending to realize spatial-temporal compositions for content re-purposing. For example, we demonstrate positioning varied manifestations of the same performance at different 3D locations with different timing, adjusting color/texture of the performer's clothing, casting spotlight shadows and synthesizing distance falloff lighting, etc, all at an interactive speed. We further develop a hybrid neural-rasterization rendering framework to support consumer-level VR headsets so that the aforementioned volumetric video viewing and editing, for the first time, can be conducted immersively in virtual 3D space.
		
	\end{abstract}

	%   <concept>
	%       <concept_id>10010147.10010371.10010382.10010236</concept_id>
	%       <concept_desc>Computing methodologies~Computational photography</concept_desc>
	%       <concept_significance>500</concept_significance>
	%       </concept>
	%   <concept>
	%       <concept_id>10010147.10010371.10010382.10010385</concept_id>
	%       <concept_desc>Computing methodologies~Image-based rendering</concept_desc>
	%       <concept_significance>500</concept_significance>
	%       </concept>
	%  </ccs2012>
	
	


	\begin{teaserfigure}
		\centering
		\includegraphics[width=1.0\linewidth]{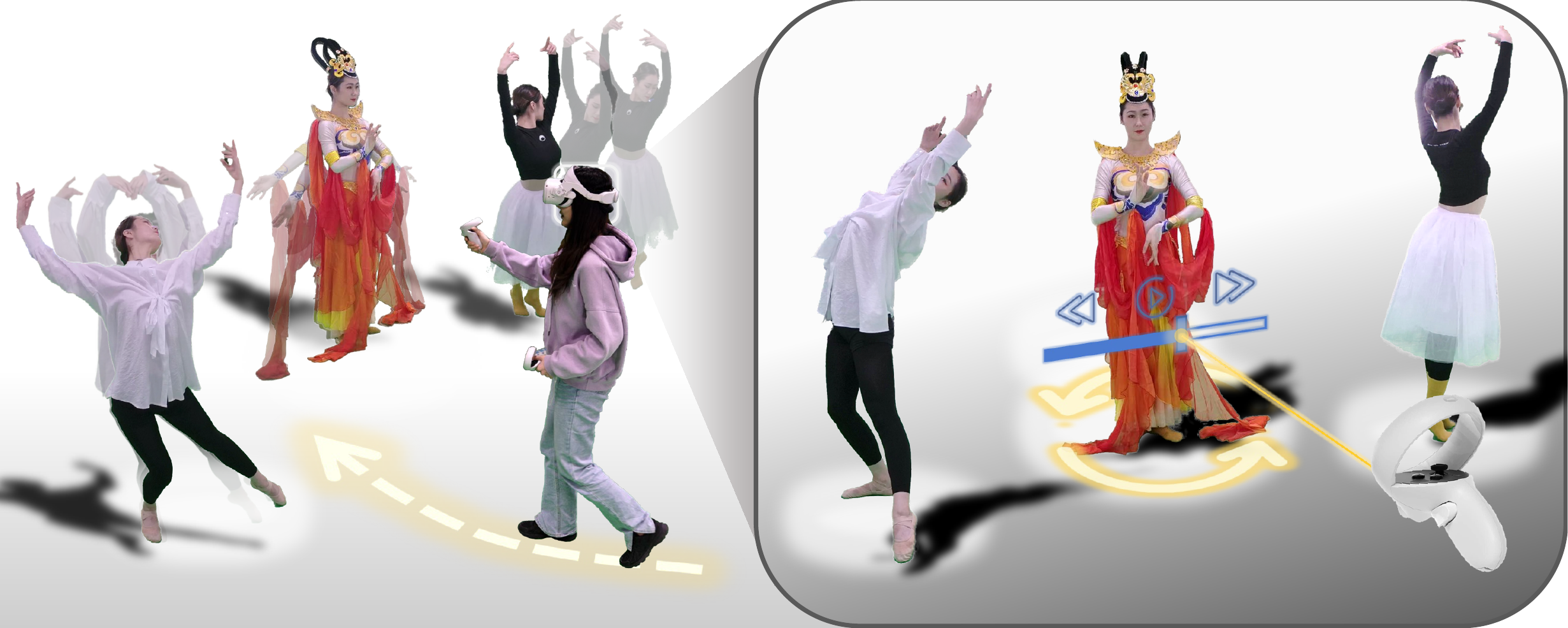}
		\vspace{-20pt}
		\caption{Our neural volumetric video technique, NeuVV, supports immersive, interactive, and spatial-temporal rendering of volumetric performances with photo-realism and in real-time. Using a hybrid neural-volumetric representation, NeuVV enables a user to move freely in 3D space to watch a single or multiple performances (Left) with a VR headset. She can also re-arrange and re-purpose the contents by adjusting the position, size, timing, and appearance of individual performers as well as adding shadow and certain lighting effects, all in real-time (right). Please refer to the supplementary video for a live recording of the experience.}
		\label{fig:teaser}
	\end{teaserfigure}

	\maketitle

	\section{Introduction}\label{sec:intro}
	Volumetric Videos (VVs), as an emerging type of visual media, are quickly expanding (not advancing) the horizons of the entertainment and movie industries with unprecedented immersive experiences. Often seen in such science-fiction films as Star Wars, VVs of human performances allow a user to move about and interact with the 3D contents with six degrees of freedom. Over the past decade, various versions of capture stages have been made available worldwide to acquire synchronized multi-view 3D videos, ranging from pure RGB camera-based systems (e.g., the CMU Panoptic Studio using hundreds of cameras~\cite{TotalCapture}) to RGBD based 3D scans (e.g., the Microsoft MR Capture Studio~\cite{collet2015high}). Yet, to provide convincing VV experiences,  volumography techniques require using a wide range of tools beyond 3D capture; they include compression, streaming,  playback, and editing.
	
	By far, the most widely adopted workflow to produce volumography is to create a dynamic mesh of the performance where each frame corresponds to a mesh with a texture map and all meshes maintain the same topology. For real performances, producing high-quality meshes imposes significant challenges: both photogrammetry and 3D scanning based reconstructions are sensitive to occlusions, lack of textures, dark textures of clothing, etc, and their results can contain holes and noises. Fixing the initial capture to meet minimal immersive viewing requirements demands excessive fixing and cleanup works by artists. A compromise is to start with a cleaned static mesh as a base and augment it with performance capture. This, however, yields to infidelity as the rigged performance appears artificial and fails to convey the nuance of the movements. Colored point cloud sequences have emerged as an alternative to meshes with a higher spatial resolution. However, they also incur much higher data rates and require specialized rendering hardware to mitigate visual artifacts.
	
	Recent advances in neural rendering ~\cite{NeuralVolumes, Wu_2020_CVPR, mildenhall2020nerf} can synthesize photo-realistic novel views without heavy reliance on geometry proxy or tedious manual labor, showing unique potentials to replace 3D capture. Most notably, the Neural Radiance Field (NeRF) ~\cite{mildenhall2020nerf} replaces the traditional notion of geometry and appearance with a single neural network where any new camera views can be realistically rendered by querying respective rays from the camera via neural inference. Despite its effectiveness, NeRF and its extensions have been largely focused on the static object, with a few exceptions ~\cite{ST-NeRF,lu2020layered, Layer-editing} to directly tackle dynamic scenes. Further, existing solutions are still a few orders of magnitudes slower than real-time to support immersive volumography. Let alone interactive, immersive content editing.
	
	In this paper, we present a new neural volumography technique, NeuVV, to push the envelope of neural rendering to tackle volumetric videos. In a nutshell, NeuVV supports real-time volumographic rendering for immersive experiences, i.e., users can view the contents in virtual 3D space and freely change viewpoints by moving around. NeuVV further provides tools for flexibly composing multiple performances in 3D space, enabling interactive editing in both spatial and temporal dimensions, and rendering a new class of volumetric special effects with high photo-realism (see Fig.~\ref{fig:teaser}). 
	The core of NeuVV is to efficiently encode a dynamic NeRF to account for appearance, geometry, and motion from all viewpoints. Analogous to 5D NeRF, dynamic NeRF maps a 6D vector (3D position + 2D view direction + 1D time) to color and density. 
	To account for angular and temporal variations at each position, i.e., view-dependent appearance, we adopt factorization schemes by hyperspherical harmonics (HH) ~\cite{avery2012hyperspherical}. 
	%Add: why?
	Further, we treat the position-specific density separately as it only exhibits temporal variations while being invariant to view directions. Hence, we further develop a learnable basis representation for temporal compaction of densities.
	The factorized color and density can be easily integrated into existing acceleration data structures such as the PlenOctree ~\cite{yu2021plenoctrees, yu2021plenoxels}. The resulting 104-dimensional vector can effectively model variations in density and view-dependent color at respective voxels. Compared to the brute-force approach of constructing per-frame PlenOctree, NeuVV tackles each volumetric video sequence as a whole, in both training and rendering, and therefore reduces the memory overhead and computational time by two orders of magnitudes. 

	By treating each dynamic NeRF as a separate entity, NeuVV supports easy spatial-temporal compositions for re-purposing the contents, and thereby immersive and interactive real-time content editing. These include real-time adjustments of the 3D locates and scales of multiple performers, re-timing and thus coordinating the performers, and even duplicating the same performer to produce varied manifestations in space and time. In addition, the employment of HHs enables temporally coherent appearance and shading editing at the voxel level. For example, we demonstrate adjusting the color/texture of the clothing, casting spotlight shadows, synthesizing distance lighting falloffs, etc, all with temporal coherence and in real-time. We further develop a hybrid neural-rasterization rendering framework that supports consumer-level head-mounted displays so that viewing and editing NeuVVs can be conducted immersively in virtual space. As a byproduct, NeuVV directly supports free-viewpoint video production at interactive speeds, enabling expert videographers to deploy their skill set on 2D video footage to volumetric videos, in a 3D virtual environment.

	To summarize, our main contributions include:
	\begin{itemize} 
		\setlength\itemsep{0em}
		\item We present a novel neural volumetric video (NeuVV) production pipeline for enabling immersive viewing and real-time interacting and editing volumetric human performances with high photo-realism.
		
		\item NeuVV employs dynamic NeRF to represent volumetric videos and adopts the hyper-spherical harmonics (HH) based Video Octree (VOctree) data structure for efficient training and rendering. 
		
		\item NeuVV further provides a broad range of composition and editing tools to support content re-arrangement and re-purposing in both space and time.
		
		\item NeuVV supports hybrid neural-rasterization rendering on consumer-level HMDs, enabling not only immersive viewing but also immersive content editing in 3D virtual environments.
	\end{itemize} 
	
	\section{RELATED WORK}
	%--------------------------------------------------------------

	%--------------------------------------------------------------
	
	%XH:
	%--------------------------------------------------------------
	%\paragraph{Volumetric videos.}
	\paragraph{Volumetric Videos.}
	%\zhang{XXX: fusion-based technique, see our editable fvv}
	%Volumetric video, free-viewpoint video or 4D reconstruction refer to the process of reconstructing 3D content over time using a multi-view setup or a monocular camera.
	Volumetric videos refer to the technique of capturing the 3D space and subsequentially viewing it on a screen.
	A volumetric video appears like a video and can be played back and viewed from a continuous range of viewpoints chosen at any time.
	A number of techniques have been proposed to synthesize point- and surface-based free-viewpoint video (FVV), including shape from silhouettes~\cite{wu2011shading,ahmed2008dense}, freeform 3D reconstruction~\cite{liu2009point,vlasic2009dynamic}
	and deformable models~\cite{carranza2003free}.

	%Fusion
	To get rid of template priors and achieve convenient deployment, one or more depth sensors can be employed to help the reconstruction.
	\cite{newcombe2015dynamicfusion} proposes a template-free real-time dynamic 3D reconstruction system. 
	Other approaches enforces the deformation field to be approximately a Killing vector field~\cite{slavcheva2017killingfusion} or a gradient flow in Sobolev space\cite{slavcheva2018sobolevfusion}.
	Pirors like skeleton~\cite{yu2017bodyfusion}, parametric body shape~\cite{yu2018doublefusion} or inertial measurement units \cite{zheng2018hybridfusion} are used to facilitate the fusion. \cite{bozic2020deepdeform} applies data-driven approaches for non-rigid 3D reconstruction.
	Rather than using a strict photometric consistency criterion, \cite{NeuralVolumes} learn a generative model that tries to best match the input images without assuming that objects in the scene are compositions of flat surfaces.
	\cite{seitz1999photorealistic,kutulakos2000theory}~recovers the occupancy and color in a voxel grid from multi-view images by evaluating the photo-consistency of each voxel in a particular order. 
	These approaches generally are difficult to tackle self occluded and textureless regions, while other approaches rely on parametric human models, which is limited to human body with tight clothes .
	In addition, they struggle with thin structures and dense semi-transparent materials (e.g., hair and smoke).

	%--------------------------------------------------------------
	\paragraph{Neural rendering.}
	
	Synthesizing photo-realistic images and videos is one of the fundamental tasks in computer vision with many applications. Traditional methods rely on explicit geometric representations, such as depth maps, point-cloud, meshes, or multi-plane images. Recently, neural rendering techniques have been showing great success in view synthesis of static or dynamic scenes with neural representations, \cite{10.1145/3450508.3464573} gives a great summary of recent work. 
	% Static
	Notably, NeRF~\cite{mildenhall2020nerf} optimizes neural radiance fields which represent each point in space with view-dependent color and density, then traditional volume rendering is applied to render images. NeRF produces unprecedented photo-realistic results for novel views and quickly becomes a research focus.  
	Similar to NeRF, recent work uses a variety of neural representations like implicit surfaces~\cite{wang2021neus, park2019deepsdf} for a more precise geometry, but they cannot handle dynamic scenes.
	% Dynamic
	To address the dynamic scene reconstruction problem, ~\cite{ park2020deformable,pumarola2020d,li2020neural,xian2020space,tretschk2020non} learn deformation fields from monocular video and then train a NeRF at canonical space. They rely on heuristic regularizations, 2D optical flow prediction, or depth images as priors, but these works suffer from large deformations and limited viewing range. 
	\cite{park2021hypernerf} further learns deformation field and radiance field in a higher-dimensional space to tackle the topological changing problem. 
	
	\cite{zhang2021editable, li2021neural, pumarola2020d} learn deformation field from multi-view videos and optimize a radiance field in canonical space, their approach supports a larger viewing range and better rendering quality compared to previous approaches. \cite{zhang2021editable} further supports certain spatial and temporal editing functions based on dynamic layered neural representations.
	\cite{peng2021neural, zhao2021humannerf} use the parametric human model as prior to learn a dynamic radiance field for human body using sparse views as inputs.
	% Summarize
	However, such works are slow to render free-viewpoint video for dynamic scenes, it takes about 30s to 1min to render a single image at $1920 \times 1080$ on a high-end GPU for NeRF, while our approach uses a hybrid representation which can more efficiently rendering for dynamic scenes in real-time.
	
	\paragraph{Accelerating NeRF.}
	% Real-time
	There are many existing work to accelerate NeRF~\cite{liu2020neural, reiser2021kilonerf, yu2021plenoctrees, Lindell20arxiv_AutoInt, 10.1145/3450626.3459863, yu2021plenoxels, muller2022instant}.
	\cite{liu2020neural} uses a sparse octree representation with a set of voxel-bounded implicit fields and achieves 10 times faster inference speed compared with the canonical NeRF.
	\cite{reiser2021kilonerf} uses thousands of tiny MLPs to speed up NeRF by more than 2000 times.
	\cite{yu2021plenoctrees} represents the view dependent colors with spherical harmonics coefficients, and extract them from a radiance field into a sparse octree-based representation, i.e., namely PlenOctree. Such representation runs 3000 times faster during rendering.
	Recently, \cite{yu2021plenoxels} directly optimizes a sparse 3D grid without any neural networks and achieves more than 100 times faster training speed up and also support real-time rendering.
	\cite{muller2022instant} achieves near-instant training time (around 5s to 1min) of neural graphics primitives with a multi-resolution hash encoding.
	Though these works are very effective at speeding up NeRF, they only support static scenes. Directly extending them to dynamic scenes suffers from expensive requirements of storage and GPU memory. Our approach uses hyperspherical harmonics and low dimensional coefficients to reduce hardware requirement, and achieves real-time inference speed.
	
	\begin{figure*}
		\includegraphics[width=\linewidth]{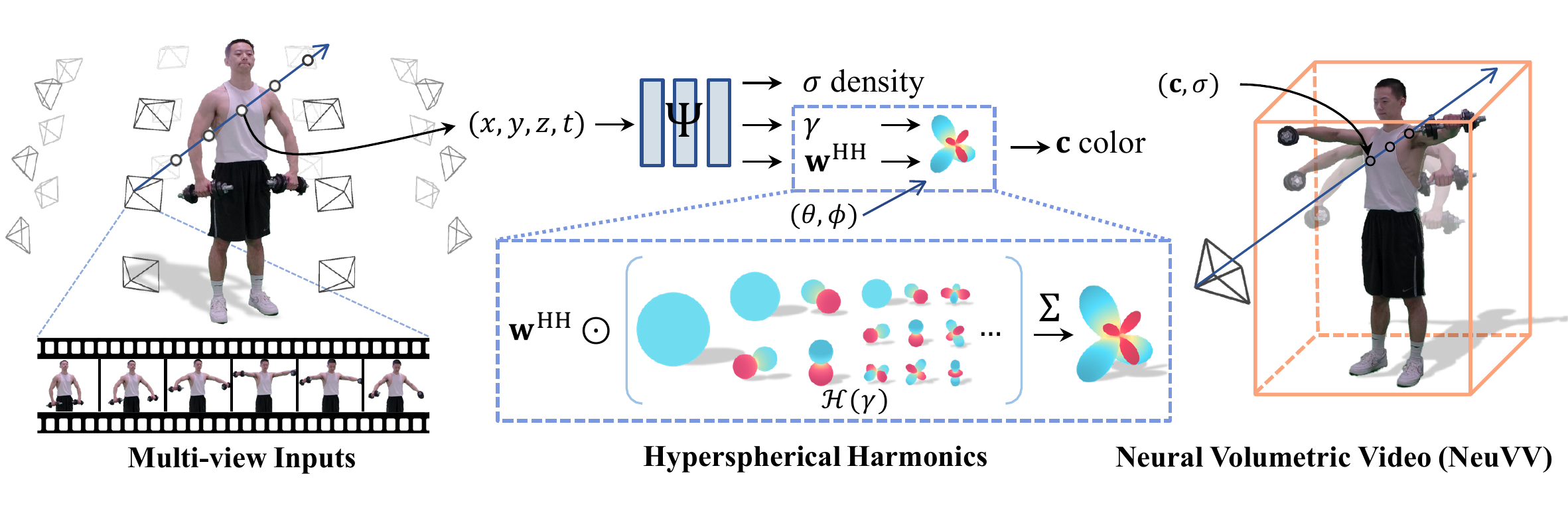}
		\caption{
			\textbf{The pipeline of our approach for neural volumetric video (NeuVV) generation.} Given a dense set of synchronized RGB videos as inputs,
			%covering a semi-circle view range
			our approach first samples a 4D points $(x,y,z,t)$ in the volumetric video, and then uses an MLP-based neural module $\Psi$ to predict density $\sigma$ and view-dependent color $c$. Instead of directly inferring color, $\Psi$ predicts coefficients $\mathbf{w}^{HH}$ of Hyperspherical Harmonics (HH) bases. $\Psi$ also predicts a hyper angle $\gamma$ which slices 4D HH into 3D Spherical Harmonics (SH), which models the view-dependent color at a specific time frame.
			%direction $(\theta, \phi)$ to 3D SH, 
			We can finally obtain color $\mathbf{c}$ and density $\sigma$ from 3D SH given the query point and ray's direction.
			Our NeuVV presents a novel neural representation of volumetric videos, which supports real-time rendering and editing of the dynamic scene when converted to a Video Octree (VOctree) representation.}
		\label{fig:pipeline}
	\end{figure*}
	
	\paragraph{Immersive Experience.}
	
	With the rapid advancement in VR/AR industry, especially with the emergence of many commercial headsets, such as Oculus Quest 2~\cite{oculus} and HTC Vive Pro 2~\cite{vive}, immersive experience is now immediately available general users.
	%
	%including the emergence of many , it has become trendy to give users an immersive experience. It requires a real-time and photo-realistic rendering technique to support its high realism. There are some headsets , e.g., Oculus Quest 2~\cite{oculus}, providing an immersive experience.
	%
	However, compared to the advance in hardware, immersive content is relative limited. Many researchers/institutes creates various devices to capture AR/VR content from the real-time, examples include the Google Jump~\cite{anderson2016jump}, Insta360 One X2~\cite{insta}), and etc for high quality 360-degree capturing.
	\cite{10.1145/3415255.3422884} proposes an approach to quickly capture high quality panorama for watching in VR headsets.
	But such panorama videos cannot support the changing of viewing location.
	%
	%Such problem can be addressed by using additional cameras and sensors. 
	%
	\cite{broxton2020immersive} presents a system to capture, reconstruct, and finally render high quality immersive videos using a semi-sphere camera array.
	\cite{10.1145/2984511.2984517} proposes a system that can achieve a real-time 3D reconstruction of the whole space using multi-view RGB-D camera arrays.
	% Technique
	Such capture systems rely heavily on explicit scene reconstruction algorithms, such as multi-view stereo~\cite{zitnick2004high, li2019learning, yao2018mvsnet}, light field~\cite{10.1145/3386569.3392485, levoy1996light,gortler1996lumigraph,buehler2001unstructured, sitzmann2021light}, multi-plane images (MPIs) representations\cite{mildenhall2019local, broxton2020immersive, srinivasan2019pushing} and image based rendering techniques~\cite{suo2021neuralhumanfvv, debevec1996modeling, carranza2003free,snavely2006photo}.
	\cite{zitnick2004high} uses multi-view stereo technique to estimate depth maps, then interpolate color images guided by the estimated depth images.
	\cite{li2019learning} learn human depth priors from thousands of Internet videos.
	% %
	% \cite{levoy1996light} uses a 4D light field to represent the 3D scene, novel view images can be treated as a set of rays which color can be obtained by interpolating captured ray color.
	%
	\cite{mildenhall2019local} uses multi-plane images which can represent complicated scenes by interpolating RGBA values on the planes. But it cannot support large changing of viewpoint. 
	%
	%Furthermore, such works cannot provide a photo-realistic immersive and interactive experience in VR applications.
	These approaches either reconstruct the scene geometry explicitly, or rely on image based rendering techniques. Reconstructing the scene geometry is always a difficult task, especially for occluded and textureless regions. On the other side, image based rendering technique produces images based on either pre-captured depth image or estimated depth, and suffers from flicking artifacts. Yet, our NeuVV does not rely on an  geometry explicitly and hence avoid the difficult geometry reconstruction problem.
	
	\section{OVERVIEW}
	
	Fig.~\ref{fig:pipeline} shows the overall processing pipeline of NeuVV. The input to each NeuVV is multi-view video sequences towards the performer. In our setting, we have constructed a multi-view dome with a set of 66 synchronized RGB videos. We use structure-from-motion to pre-calibrate the cameras so that all views have known camera poses. For validations, we select a specific frame and conduct static NeRF reconstruction. A number of options are available, from the original NeRF reconstruction~\cite{mildenhall2020nerf}, to accelerated Plenoxel~\cite{yu2021plenoxels}, and to the latest, extremely fast NGP~\cite{muller2022instant}. The reason to test on a static frame is to validate calibration as well as to support conduct better foreground/background segmentation for subsequent frames, to better produce NeuVV. Specifically, both Plenoxel and NGP provide interfaces to limit the reconstruction volume and we use the estimation for processing subsequent frames for NeuVV. 
	
	Recall NeuVV aims to approximate a dynamic radiance field using an implicit but continuous spatial-temporal scene representation, by separately factorizing the appearance, i.e., time-varying and view-dependent color, and density, i.e., changes due to motion. For the former, we apply Hyperspherical Harmonics (HH), originally designed for solving the Schr\"{o}dinger equation as basis functions. HH can be viewed as an elevation of Spherical Harmonic (SH) by considering an additional time dimension. For the latter, notice volume densities exhibit different temporal profiles than color: they are not view-dependent but can vary sharply over time. We hence use a learnable basis instead of HH for factorization.
	
	To process NeuVV, the brute-force approach would be to directly train a NeRF using the factorization, as in previous video-based NeRF~\cite{ST-NeRF, pumarola2020d}. Its downside is that NeRF is not readily supportive for real-time rendering, critical for video viewing. We hence exploit the PlenOctree designed for real-time rendering of static objects. Specifically, we extend PlenOctree to Video Octree (VOctree) to conduct network training and subsequent rendering based on HH and learnable factorizations. Finally, we integrate VOctree into OpenVR via a hybrid neural-rasterization renderer, for interaction and editing in immersive environments. NeuVV supports multiple VOctree instances as well as duplicated instances for special volumetric video effects. 
	
	\section{Neural Volumetric Video Factorization}\label{sec:method}
	
	Given a dense set of synchronized videos of dynamic performers with known camera poses, we represent the captured scene as a dynamic radiance field that can be modeled as a 6D function $\Phi$, which produces a volume density value $\sigma$ and color $\mathbf{c}$ for each space location $(x, y, z)$, time $t$ and view direction $(\theta, \phi)$, i.e.:
	\begin{equation}
		\Phi(x, y, z, \theta, \phi, t) = \sigma, \mathbf{c}
		\label{eq:6d_dynamic_radiance_function}
	\end{equation}
	\noindent A brute-force implementation is to recover one NeRF for each timestamp $t$ and then load individual frames. The approach suffers from several artifacts: it inherently incurs high memory consumption, slow training, and cross-frame inconsistency/flicking. Alternative approaches such as ST-NeRF~\cite{zhang2021editable}, D-NeRF~\cite{pumarola2020d}, NeuralBody~\cite{peng2021neural} and HumanNeRF~\cite{zhao2021humannerf} conduct spatial-temporal warping to map individual frames to a common canonical space so that they only need to train a single NeRF. The quality relies heavily on the accuracy of the estimated warping field; when deformation is large or the performer contains too few or too many textures, they tend to produce strong visual artifacts. 
	
	\begin{figure}[t]
		\includegraphics[width=\linewidth]{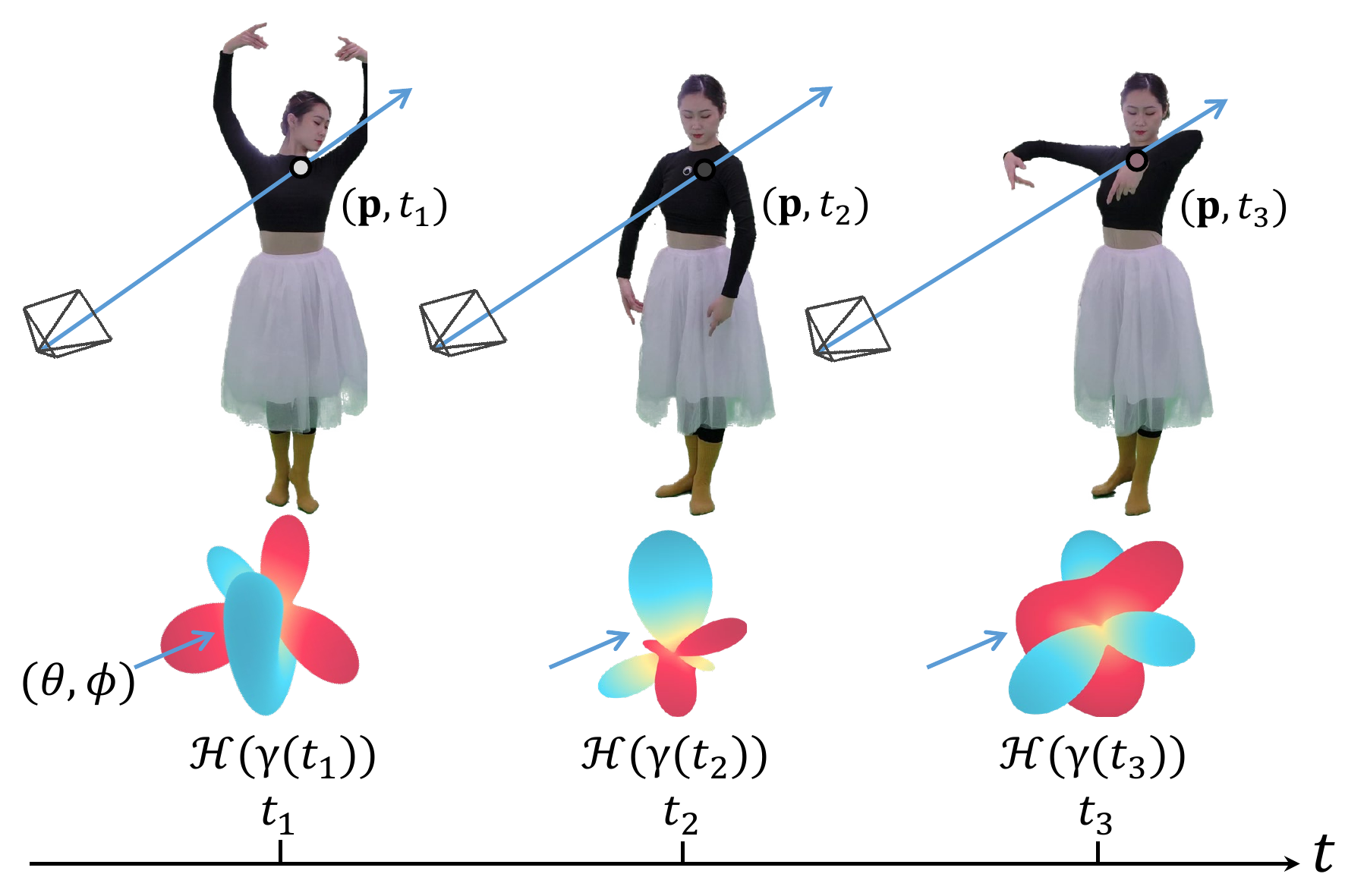}
		\caption{\textbf{Recovering color from hyperspherical harmonics.} By mapping a fixed timestamp $t$ to a hyper angle $\gamma(t)$, the 4D hyperspherical harmonics degenerates to 3D spherical harmonics. Given a spatial point $\mathbf p$ and a viewing direction $(\theta,\phi)$ along the query ray, we can recover color from spherical harmonics. }
		\label{fig:methods_hh}
	\end{figure}
	
	\subsection{Hyperspherical Harmonics Factorization}
	\label{sec:npv}
	
	NeuVV instead seeks to avoid the warping process: inspired by PlenOctree which factorizes the view-dependent appearance via spherical harmonics functions, NeuVV uses hyperspherical harmonic (HH) basis functions to further support time-variant color. Specifically, we obtain the time-varying and view-dependent color at each point $(x, y, z)$ as $\mathbf{c}(\theta, \phi, t)$ by fixing $(x, y, z)$ in Eqn.~\ref{eq:6d_dynamic_radiance_function}. 
	
	The HHs are functions of hyper angles that describe the points on a hypersphere. In the NeuVV setting, we use 4D HHs in which 3 dimensions are for describing spherical harmonics parameterized by $\theta$, $\phi$ as in PlenOctree, and 1 more dimension for the temporal dimension $t$. Consequently, we can rewrite the HH basis functions as:
	\begin{equation}\label{eqn:complex-HSH-simple}
		\mathcal{H}^m_{nl}(\theta, \phi, \gamma) = A_{n,l} \sin^l(\gamma) C^{l+1}_{n-l} \big (\cos(\gamma) \big ) \mathcal{S}^m_l(\theta, \phi)
	\end{equation}
	where 
	\begin{equation}
		A_{n,l} = (2l)!! \sqrt{\frac{2(n+1)(n-l+1)!}{\pi (n+l+1)!}}
	\end{equation}
	\noindent $\gamma \in [0,\pi]$ is the hyperangle corresponding to the time dimension, $C^{l+1}_{n-1}$ are Gengenbauer polynomials, and $\mathcal{S}^m_l$ are the 3D spherical harmonics. $l, m, n$ are integers, where $l$ denotes the degree of the HH, $m$ is the order, and $n = 0, 1, 2, ...$, following $0 \leq l \leq n$ and $-l \leq m \leq l$. Notice that when we fix $t$, HH forms an SH with a time dependent scaling factor. The complete derivations of HHs can be found in the supplementary materials.
	
	It is critical to note that all HH bases are smoothly varying function and therefore their compositions will be highly continuous and smooth in 4D space. This is preferred for view-dependent appearance, but problematic for appearance change caused by relatively fast motions at a space point. To resolve this issue, we introduce an additional a non-linear mapping function $\gamma (\cdot)$ that maps linear timestamps to hyper viewing angles, and the color then can be formulated summation of HH basis as:
	
	\begin{equation}\label{eqn:HH-color-2}
		\mathbf{c}(\theta,\phi, t) = \sum_{m,n,l}w^{m}_{nl} \mathcal{H}^m_{nl} \big ( \theta, \phi, \gamma(t) \big )
	\end{equation}
	\noindent where $w^m_{nl}$ is the coefficient of corresponding HH basis function and $\mathbf{w}^{HH}$ represents the vectorized coefficients. 
	
	\begin{figure}[t]
		\includegraphics[width=\linewidth]{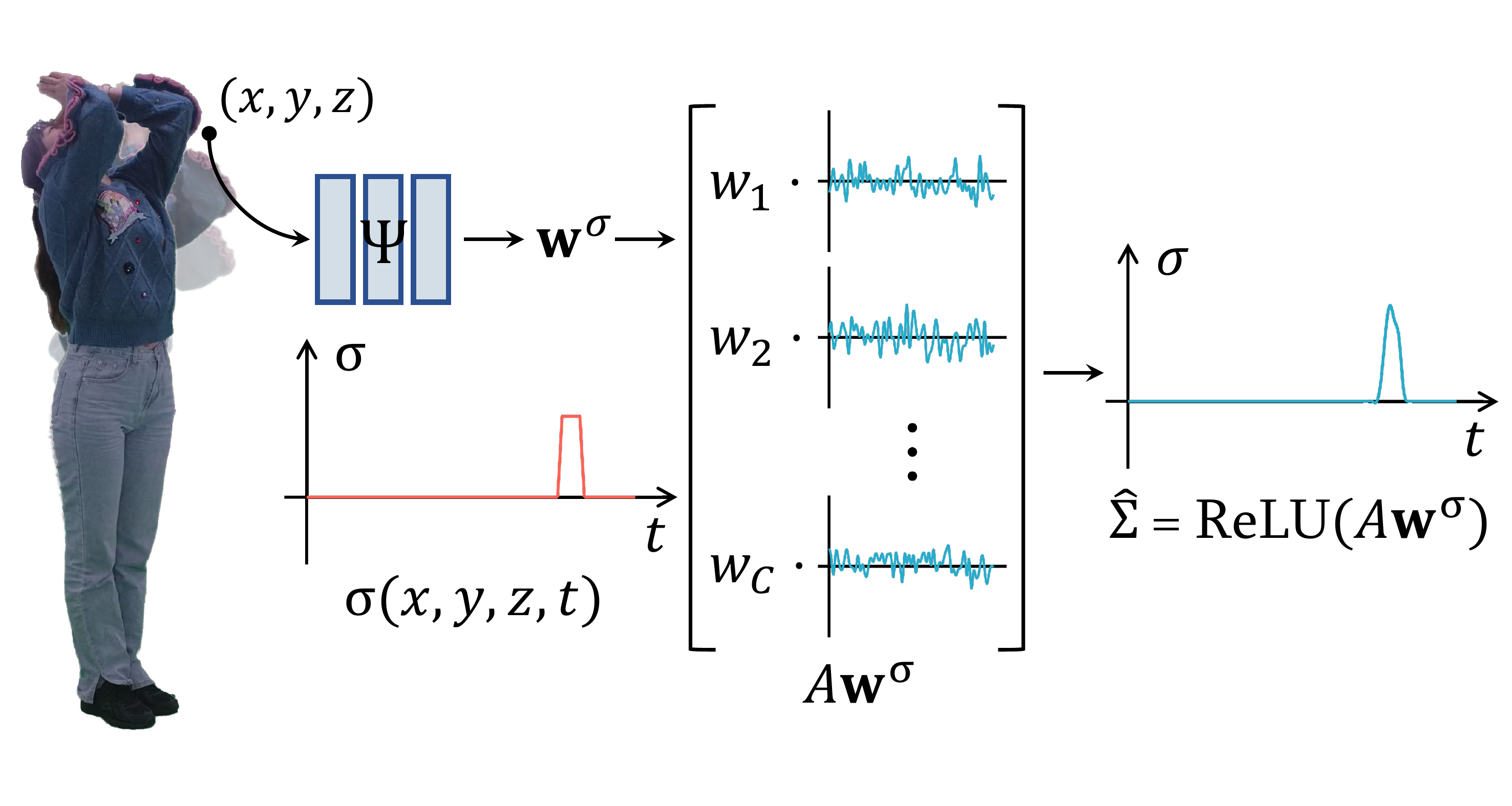}
		\caption{\textbf{Recovering density from learned bases.} For each 3D point $(x, y, z)$, time varying density can be recovered from the weighted sum of learned bases $\mathbf{w}^{\sigma}$ predicted by an MLP. As the changing of density may be rapid, we use ReLU$(\cdot)$ function for non-linear mapping and that density should not be negative. }
		\label{fig:methods_latent}
	\end{figure}
	
	\subsection{Learnable Temporal Basis Factorization}
	Once we factorize the time-varying and view-dependent color using HHs, we store volume density $\sigma(t)$ and $\gamma(t)$ for each timestamp $t$ given a spatial location.
	Notice that temporal change of volume density is caused by the occupancy/release of corresponding space point incurred by object motion. Hence temporal variations of volume density generally follow certain patterns, e.g., a moving hand passing through a space point indicates rapidly increasing from 0, staying constant, and then decreasing to 0 of the density at the point (see Fig.~\ref{fig:methods_latent} for illustration). 
	This indicates we can map the time-varying volume density onto a shared set of high dimensional bases and then use tailored low dimensional coefficients to refine the function. Such a strategy reduces memory consumption and also accelerates training. 
	
	Specifically, consider the time varying density at point $\mathbf{p}$ as $\Sigma = [\sigma_1, \sigma_2, \cdots, \sigma_N] \in \mathbb{R}^N$, where $N$ is the number of time frames. We first project it onto high dimensional density bases $A = [\mathbf{a_1}, \mathbf{a_2}, \cdots, \mathbf{a_C}]\in \mathbb{R}^{N \times C}$, where $C$ is the number of bases and $C \leq N$ for time varying density. We adopt a mapping function as:
	\begin{equation}\label{eqn:project-sigma}
		\hat{\Sigma} = \text{ReLU}(A\mathbf{w}^\sigma)
	\end{equation}
	%where $\mathbf{e}^{\sigma} \in \mathcal{N}(A)$ represents the error of the projection, $\mathcal{N}(\cdot)$ is the null space operator. 
	Same as the NeRF setting, we can use an MLP $\mathcal{P}_\sigma$ to learn the mapping weights $\mathbf{w}^\sigma \in \mathbb{R}^C$ from the spatial location $[x, y, z]$ inputs. We then optimize $A$ by minimizing the summation of differences between $\hat{\Sigma}$ and $\Sigma$ over the complete volume.
	
	Similarly, we can map the hyper angles $\Gamma = [\gamma(t_1), \gamma(t_2), \cdots, \gamma(t_N)] \in [0, \pi]^{N}$ into a set of bases $B \in \mathbb{R}^{N \times C}$ and use another MLP $\mathcal{P}_\gamma$ to estimate the mapping weights $\mathbf{w^{\gamma}}$.
	\begin{equation}\label{eqn:project-gamma}
		\hat\Gamma = \pi \cdot \text{Sigmoid}(B\mathbf{w}^\gamma)
	\end{equation}

\begin{figure}[t]
		\includegraphics[width=\linewidth]{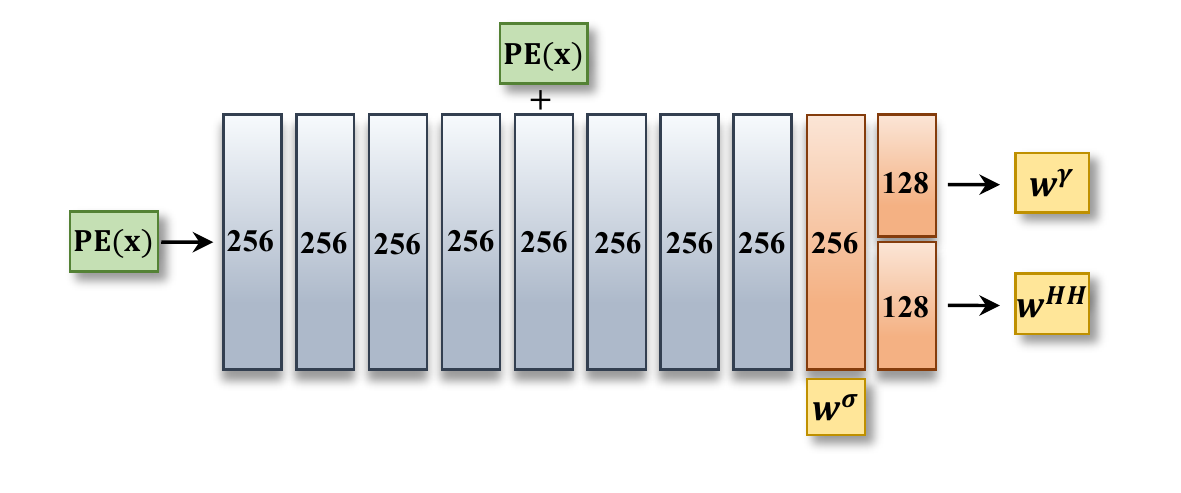}
		\caption{Details of our network structure, which basically is a multi-layer perceptron (MLP). For a 3D point $x$, we first apply positional encoding $\text{PE}(\cdot)$ and send the result to the network. The network outputs density coefficients $\mathbf{w}^{\sigma} \in \mathbb{R}^C$, hyper angle coefficients $\mathbf{w}^{\gamma} \in \mathbb{R}^L$, and HH coefficients $\mathbf{w}^{HH} \in \mathbb{R}^{L\times3}$.}
		\label{fig:network} 
	\end{figure}

	\paragraph{Neural Mapping Module.} We integrate above discussed three networks for predicting $\mathbf{w}^\sigma, \mathbf{w}^\gamma, \mathbf{w}^{HH}$ into a single MLP network $\Psi$ as illustrated in Fig.~\ref{fig:network}.
	\begin{equation}
		\Psi(x,y,z) = \mathbf{w^\sigma}, \mathbf{w^\gamma}, \mathbf{w}^{HH}
	\end{equation}
	Given a location, view direction and time tuple $(x, y, z, \theta, \phi, t)$ as input, we use $\Psi$ to predict the coefficients $\mathbf{w^\sigma}, \mathbf{w^\gamma}, \mathbf{w}^{HH}$. And we can recover the result color $\mathbf{c}$ using Eqn.~\ref{eqn:project-gamma} and~\ref{eqn:HH-color-2} and volume density $\sigma$ by Eqn.~\ref{eqn:project-sigma}.

	\paragraph{Training.} Similar to training the canonical PlenOctree for the original NeRF, we synthesize spatial-time views of NeuVV via volume rendering. Specifically, we set out to predict the color of ray $\mathbf{r}$ by sampling points along the ray and accumulate their density $\sigma_i$ and color $c_i$ as:
	\begin{equation}\label{eqn:volume-rendering}
		\begin{aligned}
			\hat{C}(\mathbf{r})= \sum_{i=1}^{|\mathcal{P}|} T_i(1-\text{exp}(-\sigma_i\delta_i))c_i\\
			\text{where }T_i = \text{exp}\left(-\sum_{j=0}^{i-1}\sigma_j\delta_j\right)
		\end{aligned}
	\end{equation}
	where $\mathcal{P} = \{p_i\}_{i=1}^{|P|}$ is the set of sampled points ordered from near to far, $\delta_i$ is the distance between the sampled points, $\exp(\cdot)$ is the exponential function.
	
	Similar to PlenOctree optimization where it is ideal to first conduct foreground/background segmentation to minimize the volume, we conduct the same foreground segmentation on NeuVV. In fact, using a video instead of an image makes automatic segmentation even easier. In our implementation, we first use the latest automatic video matting technique [VideoMatte240k] to first separate moving foreground and the static background, and then randomly select $20\%$ rays towards the background and mix them with all rays hitting the foreground to train NeuVV. We observe such a strategy is advantageous than discarding all background rays: using a small percentage of random background rays imposes additional priors to the foreground and avoids overfitting the dynamic foreground dynamic performer, especially when input views are unevenly sampled. 

	To further prevent the network from learning static background, we blend the predicted color $\hat{C}(\mathbf{r})$ with the captured background $C_{bg}(\mathbf{r})$, using weights from the predicted alpha value $\hat{\alpha}(\mathbf{r})$:
	\begin{equation}
		\hat{C'}(\mathbf{r}) = \hat{\alpha}(\mathbf{r})\cdot\hat{C}(\mathbf{r}) + (1-\hat{\alpha}(\mathbf{r}))\cdot C_{bg}(\mathbf{r})
	\end{equation}
	where $\hat{\alpha}(\mathbf{r}) = \sum_{i=1}^{|\mathcal{P}|}T_i \big(1-\exp(-\sigma_i \delta_i) \big)$. This modified rendering scheme forces the network to learn an empty space ($\sigma$ = 0) for the background part.
	
	% reconstruction loss
	Finally, we use the differences between observed colors in multi-view videos and the rendered colors from NeuVV as loss to train our model via self-supervised training:
	\begin{equation}\label{eqn:rgb-loss}
		\mathcal{L}_{rgb} = \sum_{r \in \mathcal{R}}\|C(\mathbf{r})-\hat{C'}(\mathbf{r})\|_2^2
	\end{equation}
	where $\mathcal{R}$ corresponds to the set of spatial temporal rays in each training batch and $C(\mathbf{r})$ corresponds to the captured pixel color of the input videos. We further use the same positional encoding and importance sampling scheme as in original NeRF to enhance convergence.
	
	\subsection{Video Octree (VOctree) Representation}\label{sec:octree-based-nvv}
	Same as NeRF, the brute-force approach of rendering NeuVV using MLP is slow as it requires a neural network inference for many sampling points on each query ray. For example, it takes around one minute to render a $1920\times1080$ image on NVIDIA RTX-3090 GPU, prohibitively long for deployment to real-time playback, let alone immersive rendering. We follow the PlenOctree technique [PlenOctrees] that uses a video octree (VOctree) representation with pre-tabulated density and SH coefficients for view dependent color. In our implementation, we store coefficients $\mathbf{w^\sigma}, \mathbf{w^\gamma}, \mathbf{w}^{HH}$ of each spatial location into an octree-based representation. Instead of optimizing the MLP and then tabulating the coefficients, we directly optimize the octree from the multi-view video inputs. 
	
	% 1. How to initialize PlenOctree-HH with NVV
	
	\paragraph{Initialization.} For octree based representation, its efficiency is achieved by using larger voxels for empty space while smaller voxels for occupied space with fine details. Further, ray sampling points inside the same voxel may show disturbances according to its relative position. Recall that [PlenOctree] first evaluates the density in a dense voxel grid and filter out the voxels with density lower than a threshold ($\sigma$ less than $1.0 \times 10^{-5}$), we sum up density for each voxel along time axis then filter it out use the same threshold $1.0 \times 10^{-5}$. Then inside each remaining voxel, We sample random 256 points and take the average of $\mathbf{w}^{HH}, \mathbf{w}^{\sigma}, \mathbf{w}^{\gamma}$ as the stored coefficients for the voxel. 
	
	% 2. How to Rendering
	\paragraph{Rendering.} After initialization, our VOctree based NeuVV supports rendering of dynamic entities with novel viewpoints in real-time. Specifically, given a ray we determine the voxels on its path way along with lengths of line segments $\{\delta_i\}_{i=1}^D$ inside voxels, where $i$ is the voxel index and $D$ is the total number of voxels on the ray. We fetch coefficients $\{\mathbf{w}^\sigma_i, \mathbf{w}^\gamma_i, \mathbf{w}^{HH}_i\}_{i=1}^D$ stored in the voxels. From Eqn.~\ref{eqn:project-sigma},~\ref{eqn:project-gamma},~\ref{eqn:HH-color-2}, we obtain $\{\sigma_i,c_i\}_{i=1}^D$ which are recovered density and color from coefficients, then we obtain the resultant color by volume rendering technique (Eqn.~\ref{eqn:volume-rendering})\\
	% 3. How to optimize PlenOctree-HH
	\paragraph{Optimization.} Recall the volume rendering process is differentiable. We can therefore optimize weights stored in VOctree by gradient decent using classic optimizers, such as SGD or Adam, using the RGB loss in Eqn.~\ref{eqn:rgb-loss}. For implementation, we deduct the derivatives and write custom CUDA kernels and achieve higher convergence speed, which is approximately 1,000 times faster than the original PlenOctree implementation. 
	
	Directly optimizing the VOctree, however, leads to overfitting and subsequently incurs noisy pixels on input/training video frames. We hence impose an additional regularization term to mitigate the problem. Specifically, we enforce the gradient of the difference between rendered image $\hat{I}$ and ground truth image $I$ to be small as:
	\begin{equation}
		\mathcal{L}_{grad} = \sum_{i=1}^{N\times M} \| \nabla |I_i - \hat{I_i}| \|_2^2
	\end{equation}
	where $N\times M$ is the total number of training views and $\nabla$ calculates the gradient. The final loss becomes:
	\begin{equation}
		\mathcal{L}_{total} = \mathcal{L}_{rgb} + \lambda_{grad}\mathcal{L}_{grad}
	\end{equation}
	where $\lambda_{grad}$ is a hyper-parameter to balance the RGB loss and gradient loss. In all our experiments, we set $\lambda_{grad}$ to 0.1, although it can be fine-tuned to achieve even better performances for individual datasets.
	
	\section{Immersive Rendering and Editing}\label{sec:immersive-editing}
	Existing volumetric videos have been largely used to create 2D free-viewpoint videos (FVV) \cite{wu2011shading,ahmed2008dense} where expert videographers apply their 2D footage editing skill sets. The capabilities of directly editing volumetric videos in 3D are long time dreams for content producers. The experience should be fun and compelling, with sample applications ranging from 3D visual art creations, to virtual fitness training, and to cultural heritage. Recent neural network based techniques\cite{zhang2021editable} can potentially support multi-view content editing but the process is still conducted on 2D screens rather than in 3D environment. The challenges are two-fold: 1) there lacks an immersive composition and editing tools to pair with existing VR rendering engines and headsets and 2) it is essential to achieve real-time rendering to make the 3D editing processing plausible. Since NeuVV already addresses the second challenge, we set out to design truly immersive composition and editing functionalities. 
	
	By using the VOctree to store space-time coefficients of NeuVVs, we develop a toolkit to support a variety of editing functions including spatial and temporal compositions for content re-pursing, content re-timing, and duplication and varies manifestations. Further, the Octree-based NeuVV representation enables volumetric appearance editing, e.g., we can change the color/texture of the 3D clothing worn by the performer, producing spotlight cast shadows and other relighting effects, all on the implicit representation without the need of converting the Octree to meshes. In addition, a viewer wearing the VR headset can perform along with the NeuVV where commodity motion capture solutions can be used to compare/match the move of the viewer with the virtual performer, enabling exciting new applications such as virtual fitness trainer.

	\begin{figure}[t]\label{fig:st-composition}
		\includegraphics[width=\linewidth]{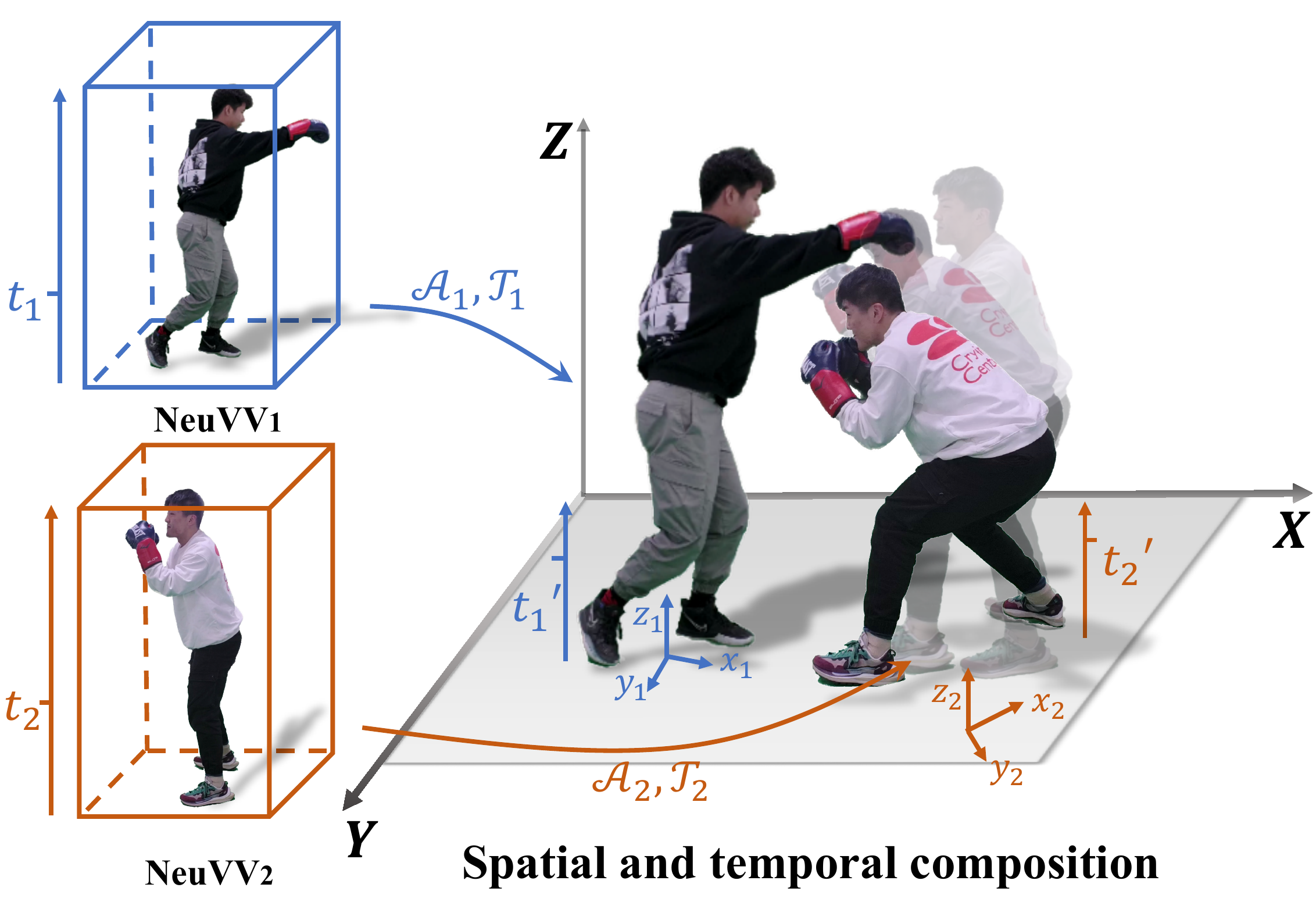}
		\caption{\textbf{Spatial and temporal composition results.} We can composite multiple NeuVVs together by applying spatial editing function $\mathcal{A}$ and temporal editing function $\mathcal{T}$. Furthermore, we use a depth-aware alpha blending strategy to generate the correct occlusion effects.}
	\end{figure}
	
	\begin{figure}[thb]
		\includegraphics[width=\linewidth]{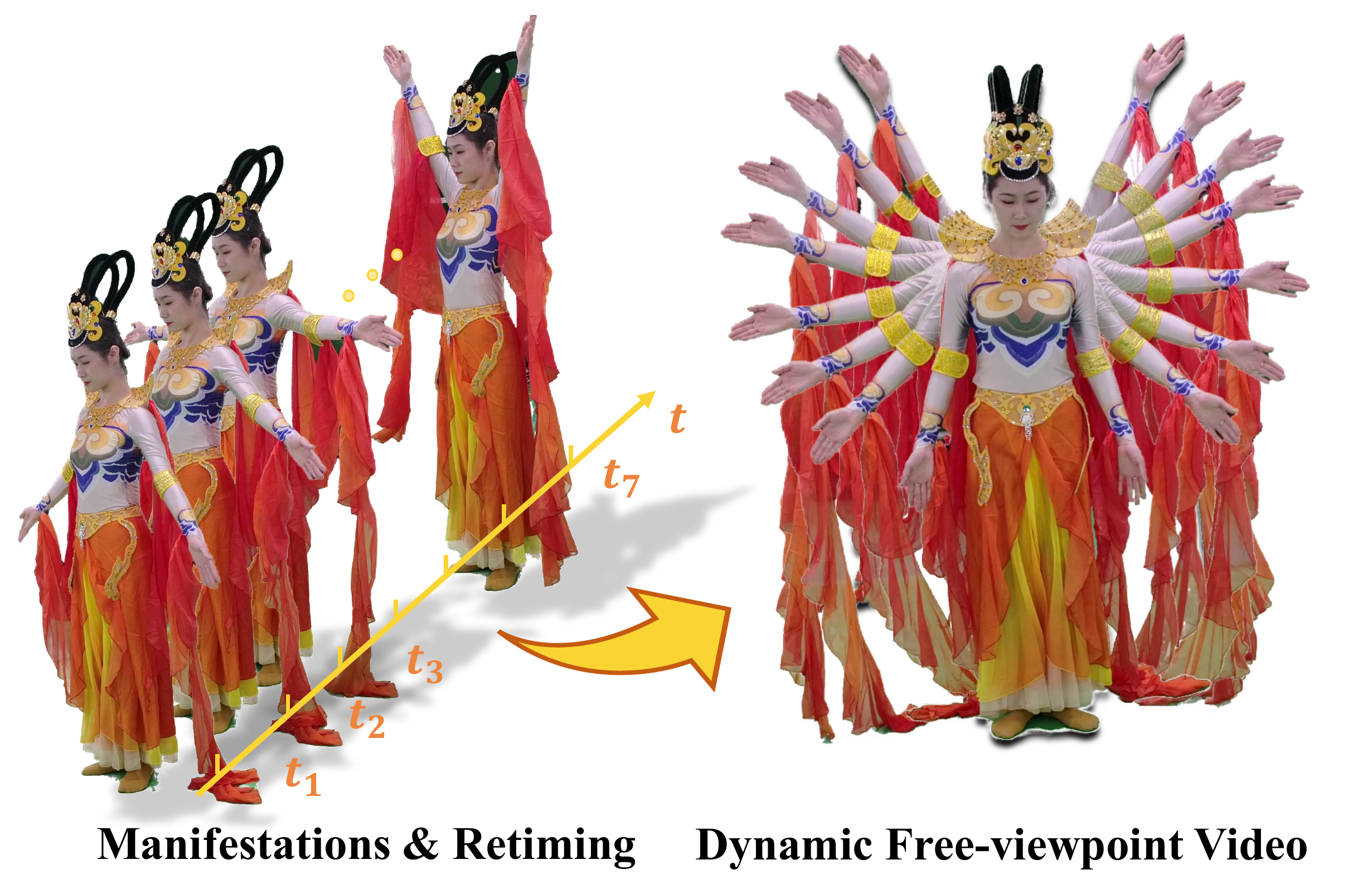}
		\caption{\textbf{Varied manifestations effect.} NeuVV achieves varied manifestations effect, \textit{Avalokiteshvara} in Buddhist mythology, using constant memory and in real-time. }
		\label{fig:manifestations}
	\end{figure}
	
	\subsection{Spatial and Temporal Composition}\label{sec:st-composition}
	
	NeuVV supports a variety of immersive spatial temporal editing operations. For spatial editing, we use the 3D bounding of NeuVV as an anchor. A user can adjust the bounding box in virtual space to scale, rotate, and re-position and re-time the performance. Since NeuVV provides a continuous representation in both space and time, the adjustment preserves photorealism in both appearance and motion. Specifically, we model the spatial editing operator in terms of an affine transformation $\mathcal{A}$. The transform from the original bounding box $\mathbf{B}$ to the adjusted one $\mathbf{B'}$ is $\mathbf{B'} = \mathcal{A}\circ\mathbf{B}$. We can subsequently apply the same transform to all nodes in the VOctree.
	
	Recall that in our NeuVV representation, the spatial and temporal dimensions are disentangled. Hence we can also manipulate the time axis to create novel temporal effects, while leaving the spatial contents unaffected. Given a sequence of timestamps $T$, we apply a general mapping function $\mathcal{T}$ to obtain a new timestamp sequence $T' = \mathcal{T}\circ T$. Typical operators of $\mathcal{T}$ include all to one mapping (pausing), clipping (partial playing), reversing (playing backward), jumping (fast forwarding), looping, etc. 
	We can hence uniformly model spatial and temporal editing as:
	\begin{equation} \label{eq:st_composite_affine}
		\Phi(\mathcal{A}\circ(x,y,z,\theta,\phi), \mathcal{T}\circ t) = \sigma, c
	\end{equation}
	
	\paragraph{Varied Manifestations.} 
	One of the most unique visual effects in NeuVV is to create varied manifestations of the same performer using only a single VOctree. The effect was popularized largely by feature film The Matrix where many copies of Agent Smith were created. Traditionally the process requires constructing a dynamic 3D model of the performer, replicate the model multiple times and position individual model at different 3D locates, and use offline rendering engines to produce the final footage. The more the duplicates, the more computational and memory resources required and the slower the rendering process. By using VOctree as the primitive, we show we can achieve real-time performance with fixed memory consumption, disregarding the number of replicates. 
	
	The brute-force approach would be to load the same NeuVV multiple times for rendering. However, since a NeuVV captures the complete plenoptic function in both space and time, one can simply use a single NeuVV where its duplicates can be treated as viewing it from different viewpoints and at different time scales. Specifically, we can reuse the composition and re-timing operators in Eqn.~\ref{eq:st_composite_affine} to produce duplicated performers positioned at different 3D locates with strategically designed, asynchronized movements. In Fig.~\ref{fig:manifestations}, we show an exemplary varied manifestation effect of the \textit{Thousand Armed Avalokiteshvara}, known in Buddhism, representing boundless great compassion. We discuss its real-time implementation in Section \ref{sec:st-composition}. 
	
	\paragraph{Depth-aware Alpha Blending}
	When we compose multiple NeuVVs as primitives (even the duplicated ones) for rendering, it is critical to conduct correct depth ordering. This is particularly important as the user is expected to move around in 3D space to view the contents at different viewpoints. Incorrect occlusions will greatly affect visual realism. To tackle such a challenging problem, we propose a simple yet effective depth blending algorithm that uses rendered depth maps $\{\hat{D}\}_{i=1}^L$ and alpha mattes $\{\hat{A}\}_{i=1}^{L}$ to guided the blending of RGB images $\{\hat{I}\}_{i=1}^L$ rendered from all NeuVVs.

	Our key insight is inspired by the traditional rendering process, i.e., the z-buffer technique more specifically. We first apply transformations to each VOctree and render the corresponding depth, alpha map and color image by tracing rays from the virtual camera. Since we adopt the octree structure, the ray tracing process can be executed very efficiently and we can do the rendering in a layer-wise manner. 
	
	For varied manifestations effects, there will be multiple iterations for generating the queried time frame one at each time and then compose all time frame results together. Since we are tracing rays from the same camera for all VOctrees, we can naturally compare the depth values of each pixel to figure out the occlusion relations and then conduct canonical alpha blending without difficulty. This process is illustrated in (Algorithm.~\ref{alg:alpha-blending}).

	\begin{algorithm}[thbp]
		\SetKwInOut{Input}{Input}\SetKwInOut{Output}{Output}
		\caption{Depth-aware Alpha Blending}
		\SetAlgoLined
		\Input{$\{I_i\}_{i=1}^L, \{D_i\}_{i=1}^L, \{A_i\}_{i=1}^{L}$} 
		\textbf{Initialization:} $I = I_1, D = D_1, A = A_1$\\
		\For{$i = 2,\ldots, L$}{
			$fg = D_i <= D, bg = D_i > D$ \\
			$I[fg] = A_i[fg]I_i[fg] + (1-A_i[fg])A[fg]I[fg]$ \\
			$I[bg] = A[bg]I[bg] + (1-A[bg])A_i[bg]I_i[bg]$ \\
			$D[fg] = D_i[fg]$\\
			$A = A + A_i\cdot(1-A)$
		}
		\Output{Blended RGB image $I$, depth $D$, and alpha image $A$}
		\label{alg:alpha-blending}
	\end{algorithm}

	\subsection{Editing and Rendering}
	
	Our VOctree-based NVV representation further supports certain levels of appearance editing. Adding lighting effects or changing appearance coherently in both space and time have been particularly challenging on volumetric videos. In the 2D videos, rotoscoping is widely adopted for tracking objects over frames and subsequently consistent recoloring and retexuring. For volumetric videos, it is simply infeasible to consistent rotoscope over all frames and at all viewpoints. For NeuVV, the more challenging task is adding lighting effects: as an implicit representation, NeuVV does not produce explicit geometry such as a mesh that can be used for adding lighting effects. We demonstrate how to use the VOctree structure to achieve certain classes of appearance editing and relighting effects. 
	
	\paragraph{Appearance editing.} To edit appearance, we can first select the set of voxels of interests. If the edits are conducted on 2D screen (e.g., for FVV generation), one can use images/frames to map highlighted pixels to their corresponding voxels. Under the VR setting, they can be directly conducted in 3D space by defining a 3D region and selecting the voxels within using the controller.
	Recall, NeuVVs adopts an implicit representation with coefficients $w^\sigma$ as latent variables, direct editing of these coefficients, although possible, does not readily produce meaningful results. Our editing function therefore aims to modify the appearance of the corresponding VOctree rather than the content itself. Nonetheless, this is sufficient for the user to modify the texture and color of clothing. Specifically, we append 5 additional channels to each voxel that represent the target RGB values $\mathbf{c}_d$, the target density value $\sigma_d$, and the timestamps $t_d$ indicating which frames on this voxel should be modified. 
	
\begin{figure*}[tbp]
		\includegraphics[width=0.9\linewidth]{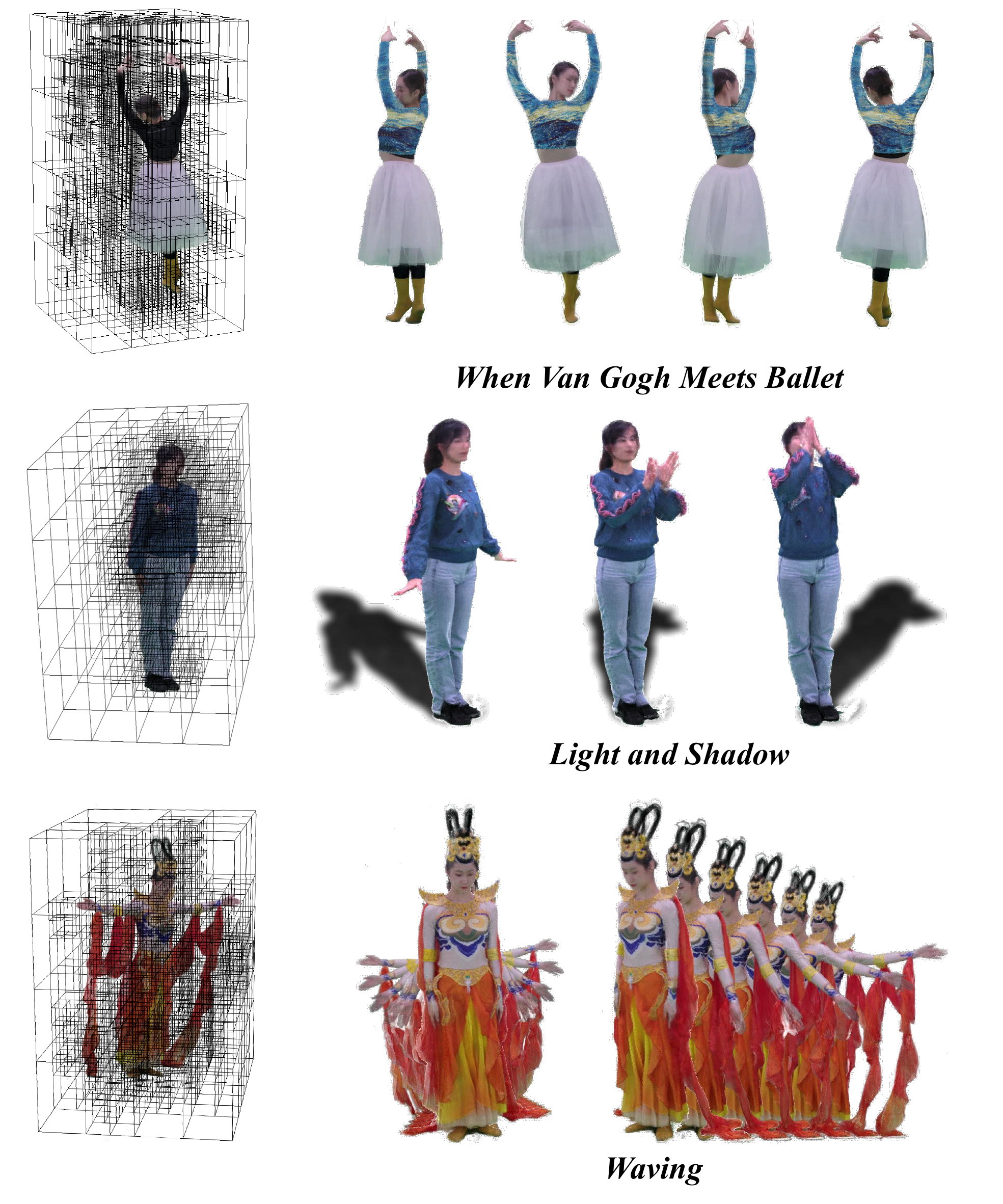}
		\caption{\textbf{Editing results.} For \textit{When Van Gogh Meets Ballet} (top), we edit the clothes appearance by mapping Van Gogh's famous painting \textit{Starry night}, and show some representative views. For \textit{Light and shadow} (middle), we add virtual light and cast the shadow of performers as virtual motion magnifier, we show representative frames of edited VOctree of performer. For \textit{Waving} (bottom), to create waving effect of the same performer, we duplicate and shift her location and timing.}
		\label{fig:gallery}
	\end{figure*}

	The challenge here is to determine which voxels to be edited and how to blend with the original NeuVV VOctree. Consider painting a 2D pattern over the NeuVV. Given the camera pose, we trace each pixel/ray towards the VOctree and we locate the terminating voxel along the ray when the accumulated alpha rendered using NeuVV is beyond a threshold (0.99 in our implementation). We then assign the target color to the voxel. At render time, the target color can be further blended with the NeuVV rendering results to further improve view-consistency. In Fig.~\ref{fig:gallery}, we show free-viewpoint rendering of a ballerina sequence after we paint Van Gogh's starry night onto the original black tight shirt.
	Note the complexity of appearance editing, via either region selection or ray tracing,  is significantly lower than the volume rendering process with HH, as it does not require volume integration. So the appearance editing is still in real-time and can be done interactively during the dynamic rendering process.
	
	\paragraph{Spotlight Lighting.} In a theatrical setting, spotlight produces artistic effects for enhancing realism. They also help convey the nuance of human motion: when motion is minute, its shadow variations can be still be highly apparent attributed to perspective magnifications. Such changes of light and shadow can increase the viewing experience of the viewer.

	Producing spotlight shadows of NeuVVs is nearly identical to rendering shadow map of meshes: we can position a virtual camera in the position of the point light source and render the VOctree at the respective viewpoint. In traditional shadow maps, shadows are created by conduct a visibility test using the z-buffer. Since NeuVV builds on top volume rendering, we further use the accumulated alpha values along rays. Specifically, we first render an alpha map from the point light virtual camera, reserve the alpha map (as the denser the alpha map, the higher the probability the ray been blocked and hence induces shadow), and finally project it onto the ground. For faster rendering, we choose to render shadows at lower resolution and then use low pass Gaussian filters to remove \textit{Moire patterns}. Fig. \ref{fig:gallery} shows sample cast shadows of a dynamic performer. The figure and the supplementary video demonstrate that under the VR setting, NeuVV produces visually consistent shadows for better conveying subtle motions.  

	Another lighting effect is distance falloff: the closer the part of the performer to the light source, the brighter it appears. Specifically, instead of using the density accumulation as in shadow maps, we directly render the depth map and compute the falloff in terms of the distance between the voxel to the light. If we position the spotlight on top of the performers, their faces will appear brighter than feet, creating special theatrical atmosphere. Under the VR setting, we observe they produce more realistic encounters for viewing volumetric performances.
	
	\subsection{2D vs. 3D Rendering}
	\begin{figure}[t]
		\includegraphics[width=\linewidth]{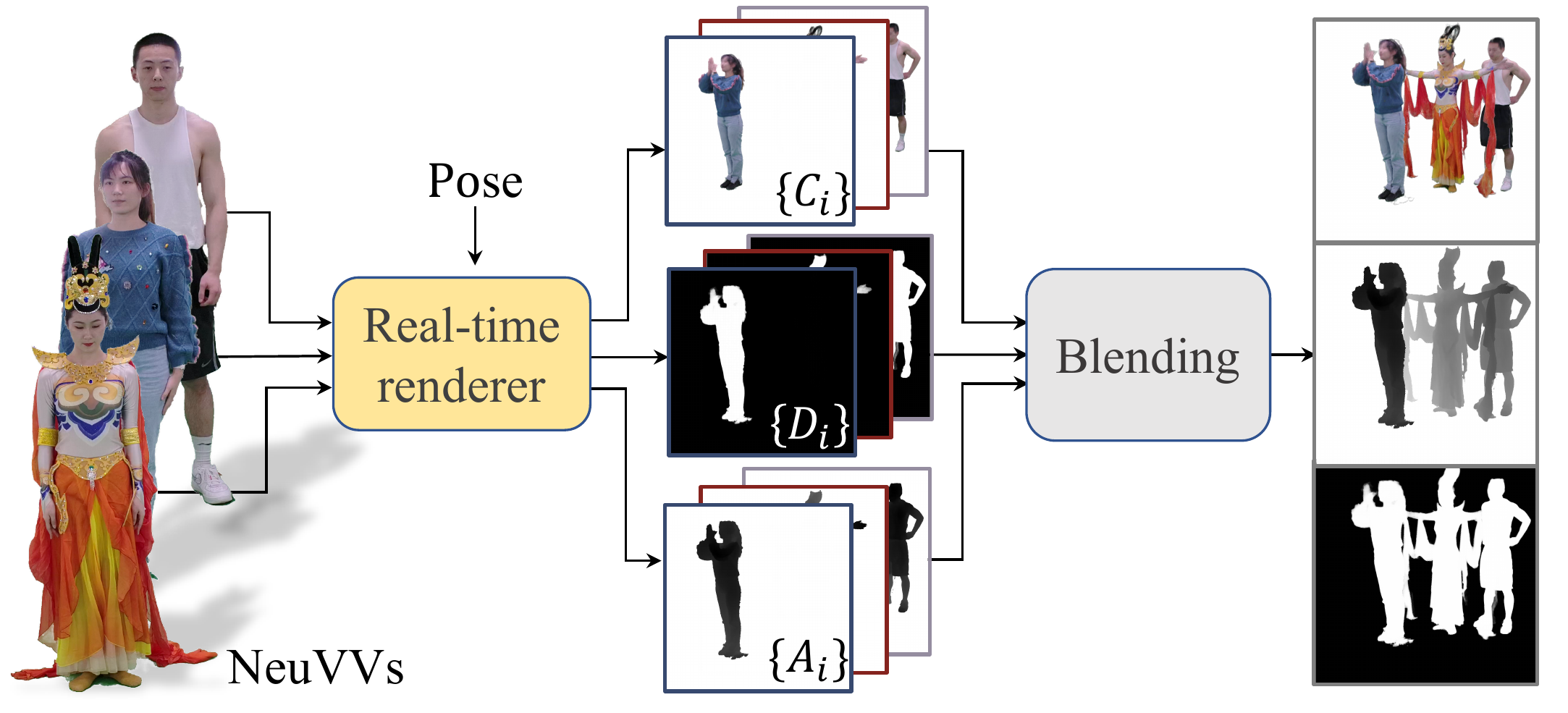}
		\caption{\textbf{NeuVV rendering under VR setting.} Our VR rendering pipeline, which renders multiple NVVs' alpha, depth and RGB images simoutanouly in real-time at given camera pose. Then we blend the images together in a depth-aware manner.}
		\label{fig:vr-pipeline} 
	\end{figure}
	% 承上
	While most volumetric videos are processed or viewed on desktops including the latest neural representations, the best viewing and editing experiences should be immersive and hence carried out under the VR setting when headsets are available. We have implemented NeuVV renderers under both settings. 
	
	\paragraph{Free-Viewpoint Video Renderer}
	We first develop a Free-viewpoint Video (FVV) renderer based on NeuVV. Most existing FVV players are based on 3D meshes or points, popularized by Microsoft Capture Studio. The use of explicit representations have its advantages and limitations: mesh rendering is directly supported by the graphic hardware and can be integrated into existing rendering engines; yet producing high quality meshes without extensive cleanup of the initial capture is still extremely difficult. NeuVV's implicit representation addresses the visual quality issue but additional efforts are needed to fit it to existing rendering pipelines. 
	
	In our implementation, VOctree builds upon the open source PlenOctree originally designed for real-time rendering of NeRF-based static objects. We modify the spherical harmonics (SH) bases in PlenOctree and replace them with our HH bases for appearance rendering and learnable bases for density and hyper angle. It is worth nothing VOctree supports the rendering a single performer and multiple performers. For the former, ultra-fast rendering at a lower resolution helps to check the quality of the trained neural representations, e.g., to determine if the spatial-temporal videos can be sufficiently replicated by the network with acceptable visual quality. For the latter, it is particularly useful for re-purposing the contents by obtaining real-time feedback on the final layout and visual effects of the FVV. This is particularly important as many previous FVV generators, including the neural ones, require long processing time instead of being interactively editable. In our implementation, we have rewritten custom CUDA kernels as well as added rendering capabilities of shadows and light falloff effects via the alpha and depth maps. Once validated, the contents can be transferred to the VR renderer to create immersive experiences. 

	\begin{figure}
		\includegraphics[width=\linewidth]{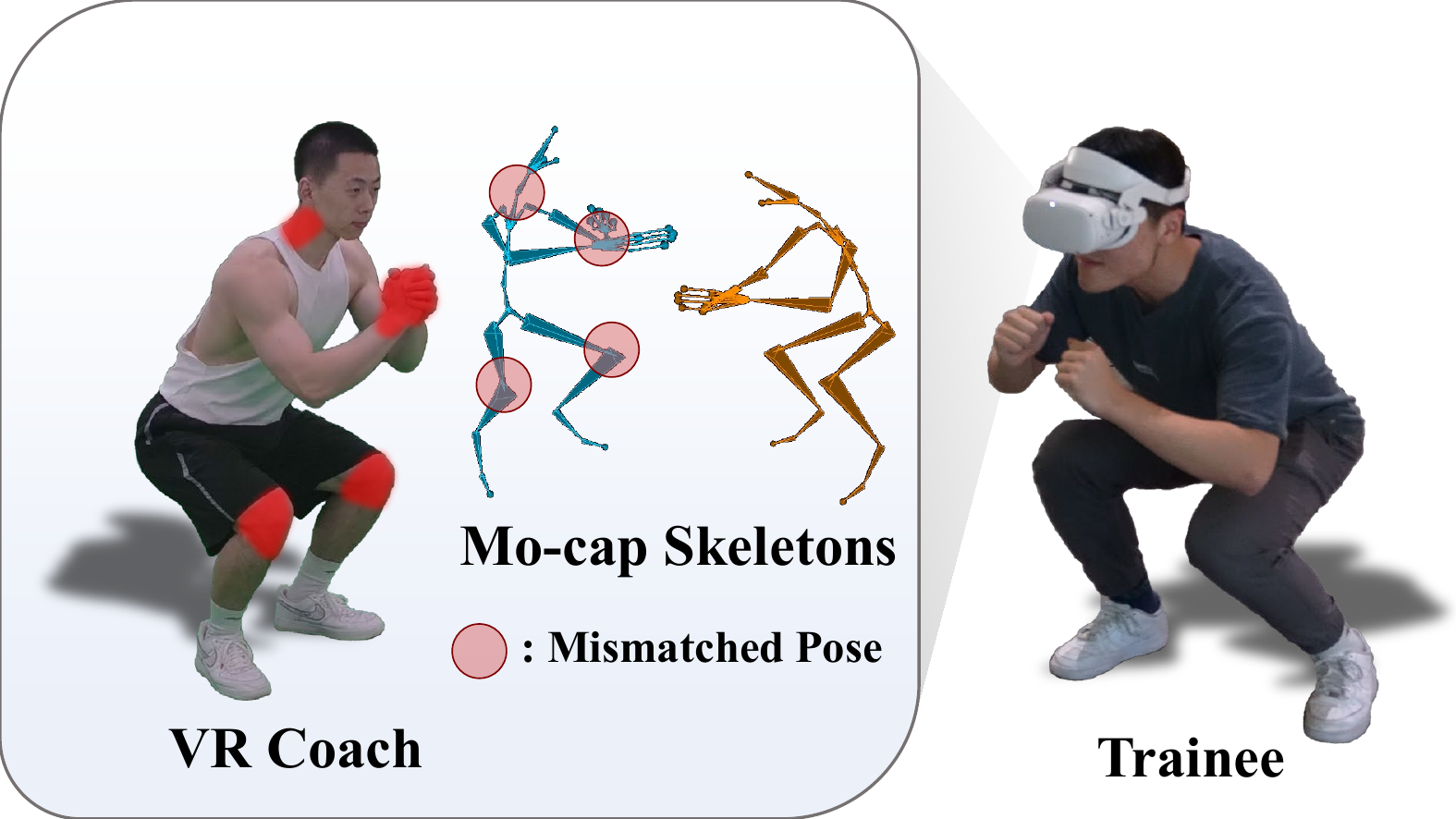}
		\caption{\textbf{Immersive fitness training demo.} Our NeuVV renders views of the coach in real-time and highlights body parts (red) corresponding to the incorrect joints based on the differences between the reference skeleton and the skeleton generated by mo-cap, which helps trainee to correct poses. }
		\label{fig:fitness} 
	\end{figure}
	% 1. Technical 2. Editing ability
	\paragraph{VR Renderer.} 
	
	Most unique to our NeuVV renderer is its support for head-mounted displays (HMDs). We have developed a NeuVV API based on OpenVR for supporting different types of HMDs (Oculus, Mixed Reality, Vive, etc). In several examples shown in the paper and the video, we demonstrate NeuVV VR rendering using Oculus Quest 2 on a single NVIDIA RTX-3090 GPU. We render stereo views at a resolution of $1920\times 1080$. The NeuVV API takes camera pose of the headset from OpenVR and renders individual VOctrees representing different performers with algorithms discussed in Section XXX to tackle correct depth ordering. Shadows and falloff lighting can be turned on and off using the controller. Fig.~\ref{fig:vr-pipeline} shows the complete NeuVV VR rendering pipeline.   
	A key advantage of the VR renderer is it allows a user to compose and edit volumetric videos in 3D space. We provide a group of interaction functionalities. For selection, we use the position and the orientation of the controller to emit a line (ray) towards the scene for selecting the target NeuVV in terms of its bounding box. Once selected, the content can be re-positioned freely in 3D space, as if a user is controlling a 3D object, largely thanks to real-time VOctree rendering. We also provide a self-rotation function where the performer self-rotates smoothly along the y-axis while the video plays along. 
	
	Recall that the original PlenOctree only supports free-viewpoint viewing, i.e, the camera pose can change but the object cannot rotate otherwise the its corresponding tree structure needs to be reconstructed. Therefore, we emulate rotation of an NeuVV by transforming the viewpoint with respective to each individual entity, i.e., we compute the corresponding viewpoint for each NeuVV within the scene. To be more specific, we make the camera rotate around the performer and keep it look at the performer.
	
	To realize duplicated manifestation, our system provides a duplicated button. Instead of making multiple copies of the VOctree which will significantly increase memory consumption, we only create a new pointer to the same VOctree, along with the transformed viewpoint and the desired re-timing map, as if it were a different NeuVV. Rendering can then be carried as usual with depth ordering support. In this way, we can create as many duplicates as possible without incurring additional memory overhead. Finally, as NeuVV can also be viewed as a video, we provide pause/play/forward/backward controls on the controller, each implemented by adjusting respective timestamp controls as shown in Section \ref{sec:st-composition}. The supplementary video provides many examples demonstrating the NeuVV VR experience.
	
	\begin{figure}[t]
		\includegraphics[width=\linewidth]{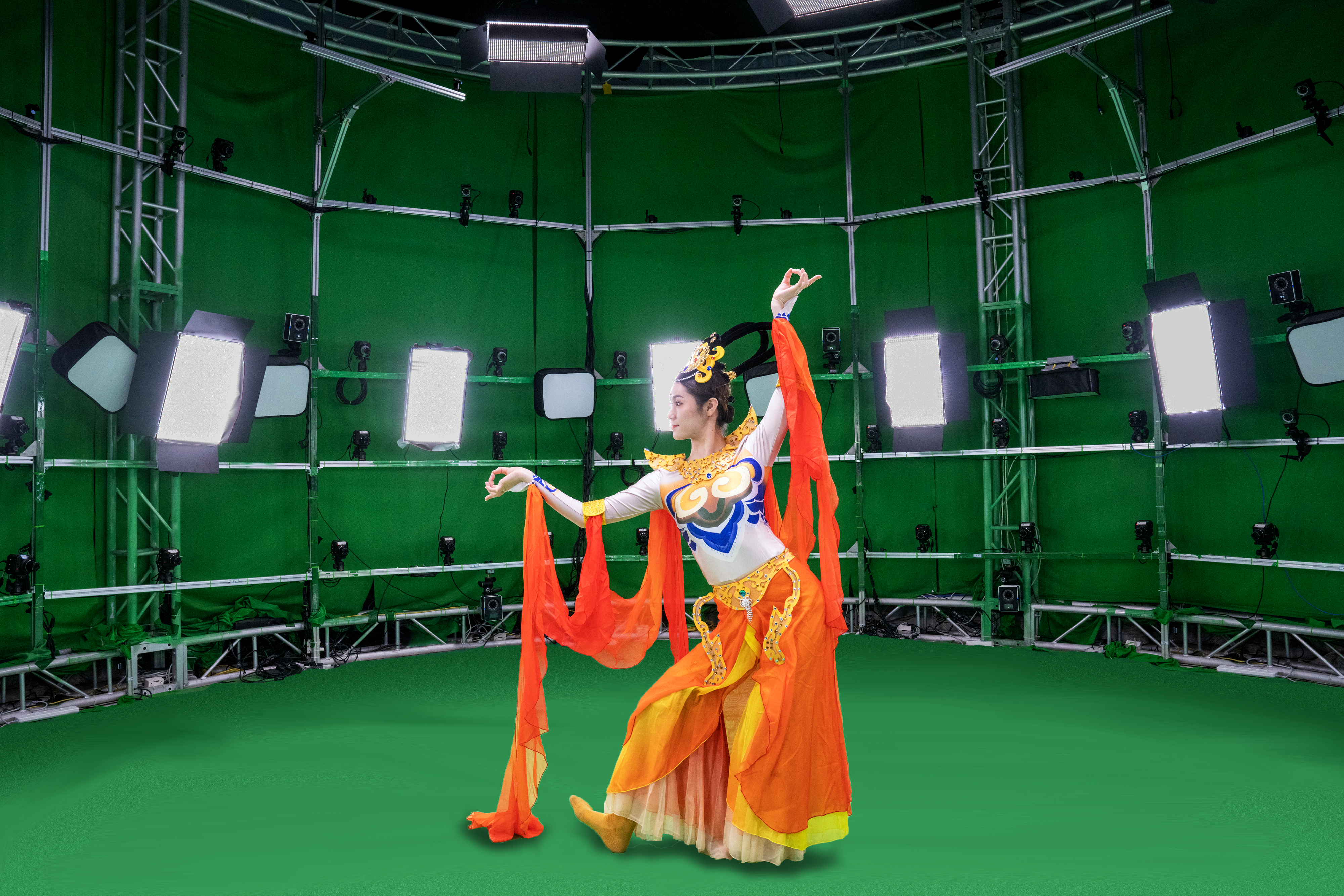}
		\caption{\textbf{Capture system}. Our capture system consists of 66 industry Z-CAM cameras which are uniformly arranged around the captured performer to cover a view range up to 1440 degrees (4 circles). Each of camera circles is focused on lower body, full body, upper body or top views of performers. All the cameras are calibrated and synchronized in advance, producing 66 RGB streams at 3840 $\times$ 2160 resolution and 25 frames per-second.}
		\label{fig:capture_system} 
	\end{figure}
	
	\paragraph{Live User Motion Feedback.} In addition to composition and editing, we allow the user to perform along with the virtual performers in NeuVV. A potentially useful function is hence to highlight live user motions on the top of the NeuVV footage. This is particularly useful for fitness training and dancing games in VR setting, i.e., a home personal training who will remind the user about incorrect postures that can also adverse effects.  
	
	There are many real-time motion capture solutions available and we adopt the recent single camera technique\cite{he2021challencap} for convenience. It is able to detect 21 key points of skeletons. We have developed an interface to our VR NeuVV to allow the estimated mo-cap results feed directly back to the renderer. As a reference, we preprocess the NeuVV of the trainer by conducting multi-view skeleton detection. Notice that many of the volumetric videos in this paper were captured using a dome system where each camera only captures a partial view of the performer and skeleton detection is less robust. Therefore we first render a multi-view full body sequence using NeuVV and then conduct skeleton extraction. This produces very high quality skeletal movements. We then compare the user movements with the performer's and highlight their differences in live viewing experiences.

	\begin{figure*}[thb]
		\includegraphics[width= 1.0 \linewidth]{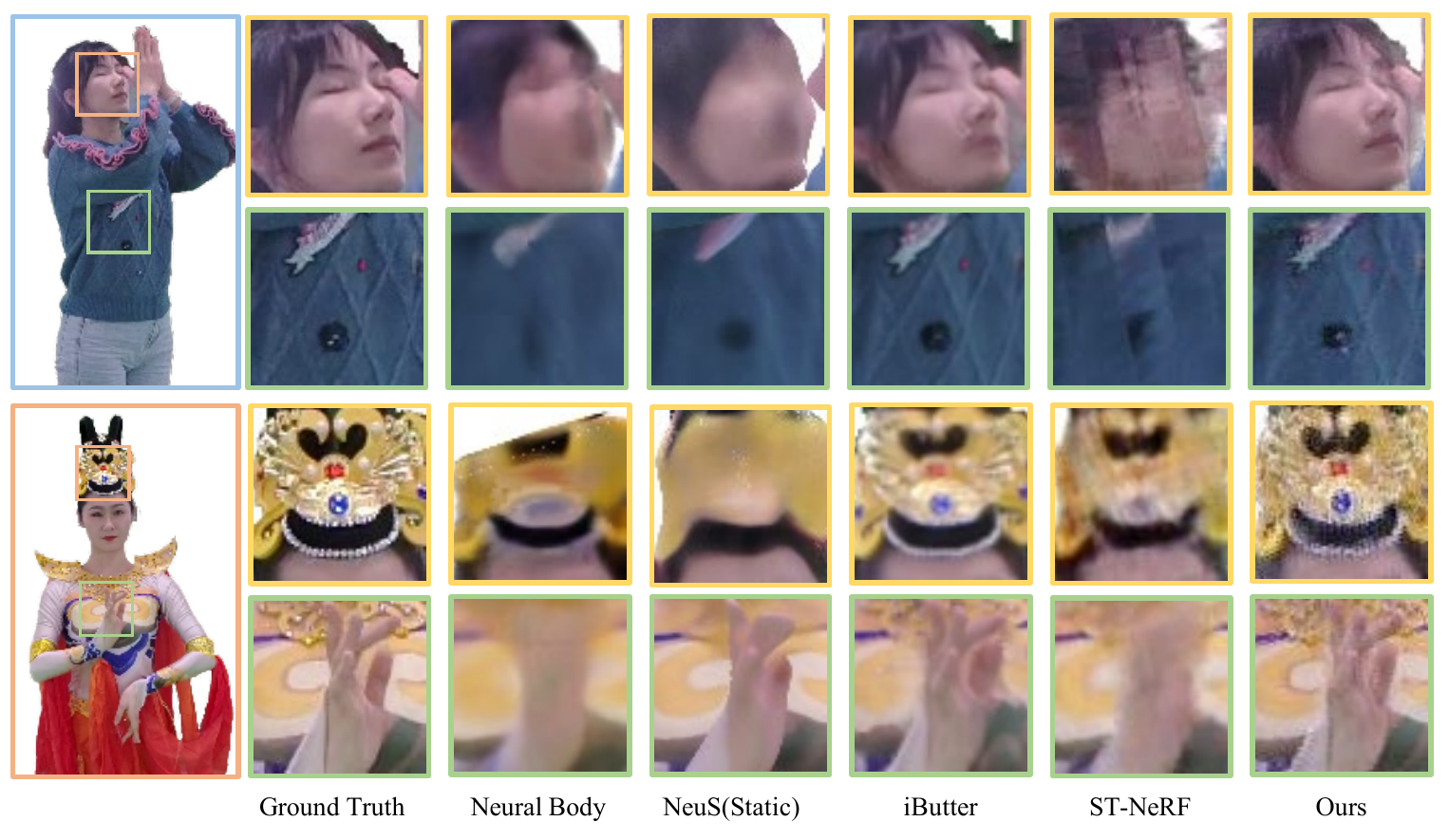}
		\caption{Qualitative comparison with Neural Volumes, Neural Body, iButter and ST-NeRF. Note that our approach generalizes more photo-realistic and finer details.}
		\label{fig:compare}
	\end{figure*}

	Fig.\ref{fig:fitness} shows a typical example of fitness training where the user conducts deep squad, one of the most important movements in leg training, along with the virtual trainer represented using NeuVV. The details of squat movements are very important for the effectiveness and safety of training, which is difficult for beginners to grasp.
	Once motion discrepencies are detected, our renderer not only highlights their differences but also suggests the user moving about the trainer to observe the correct movements from the most suitable view angle, a special treat provided by volumetric videos. Any time, the user can use the VR controllers to pause, remind, re-position, and scale the video content at will. 
	\section{RESULTS}\label{sec:results}
	
	We have validated NeuVV factorization on challenging volumetric videos captured under real-world settings as well as implemented an array of composition and editing tools suitable for 2D screens and 3D immersive environments.  We provide implementation details as well as the utilized datasets captured by our multi-view system. We further compare NeuVV vs. other alternatives, most of which are offline algorithms though. Nonetheless, we show NeuVV outperforms them in visual quality and is much faster. We also discuss different components of NeuVV and how they affect the results qualitatively and quantitatively. Finally, we illustrate spatial-temporal composition and editing functionalities of NeuVV as well as discuss its limitations. 

	\paragraph{Implementation Details.}
	We have implemented the core NeuVV component, i.e., VOctree (Section~\ref{sec:octree-based-nvv}) in PyTorch with customized CUDA kernels for inference and back propagation. All experiments are trained and optimized using a single NVIDIA Tesla A100 GPU or a NVIDIA GeForce RTX3090 GPU. Real-time rendering either on s 2D screen or VR headset is conducted on a single RTX3090. 
	The most time consuming component of NeuVV is training and generation. Depending on the number of video frames in the captured scene (75 to 150 frames) and the complexity of the performer's motion, the training time ranges from 12 to 24 hours with an input resolution of $960\times 540$, followed by a conversion from NeuVV to VOctree which takes around 15 minutes per sequence. Finally, we optimize VOctree-based NeuVV with an input image resolution $1920 \times 1080$ where the processing time ranges from 8 to 12 hours. 
	
	\paragraph{Datasets.} 
	
	We have captured ~20 multi-view video sequences, all with a single performer acting inside the capture dome. Motions range from relatively static movements such as hand waving to moderate ones as fitness training and dramatic ones as dancing. We also have the performers wearing various types of clothing, from high tight outfits as in the Ballerina sequence to high loose dresses and robes in the Dunhuang dance sequence, to test the robustness of our approach.
	Fig.~\ref{fig:capture_system} shows our capture system that consists of 66 industry Z-CAM cameras which are uniformly arranged around the performer covering a view range up to 1440 degrees (4 circles at different latitudes). All the cameras are calibrated and synchronized in advance, producing 66 RGB streams at 3840 $\times$ 2160 resolution and 25 frames per-second. 
	
	In order to obtain a high quality dataset, we have specially designed our capture system. First, to obtain more detailed acquisition images, we orient the cameras along the equator and on the second circle from top down to face the lower and upper body of the performer, respectively. Cameras on the rest two circles (the highest and the second lowest) are used to capture the complete (full body) performer within their field-of-view. This strategy helps to balance the resolution and reconstruction quality: if all views capture individual fragments of the body, the calibration process will lead to large errors and subsequently affect NeRF/NeuVV reconstruction; If all views capture full body, the final resolution on faces and clothing will be low. Our compromise ensures both high quality calibration and preservation of fine details. The numbers of frames used in NeuVV range from 75 to 150 (3s to 6s), depending on motion range and speed, in line with previous approaches.
	
	\begin{table}[t]
		\caption{\textbf{Quantitative comparison against several methods in terms of rendering accuracy.} Compared with ST-NeRF, NeuS, NeuralBody and iButter , our approach achieves the best performance in \textbf{PSNR},\textbf{SSIM} and \textbf{MAE} metrics. Note that NeuS is per-frame training.}
		\centering
		\small{
			\begin{tabular}{l|c|c|c|c | c}
				\multicolumn{5}{c}{ \colorbox{best1}{best} \colorbox{best2}{second-best} } \\
				Method        &  PSNR$\uparrow$        & SSIM$\uparrow$          &MAE$\downarrow$          &LPIPS$\downarrow$ & Realtime  \\ \hline
				Neural Body   & 29.20   & 0.9777     & 0.0068     & 0.0728 & \XSolidBrush  \\
				NueS          &27.07  & 0.9828  & 0.0053  & \cellcolor{best2}0.0410  & \XSolidBrush    \\
				iButter       & \cellcolor{best2}32.76  & \cellcolor{best2}0.9859 & 0.0609 & \cellcolor{best1}0.0032& \XSolidBrush  \\
				ST-NeRF       & 32.57 & 0.9687 & \cellcolor{best2}0.0043 & 0.0570 & \XSolidBrush  \\ \hline
				Ours          & \cellcolor{best1}34.27 & \cellcolor{best1}0.9875 & \cellcolor{best1}0.0034 & 0.0529 & \Checkmark \\    \hline
				% 		Ours & \textbf{36.43} &  \textbf{0.9969} & \textbf{0.0013} & \textbf{0.0182}\\    \hline
			\end{tabular}
		}
		\rule{0pt}{0.05pt}
		\label{table:quantitative comparison_perform}
		\vspace{-1mm}
	\end{table}

	\subsection{Rendering Comparisons}
	\paragraph{Comparisons to SOTA}
	Our approach is the first neural representation which enables real-time dynamic rendering and editing and to the best of our knowledge.
	To demonstrate the overall performance of our approach, we compare to the existing free-viewpoint video methods based on neural rendering, including the implicit methods \textbf{NeuS}~\cite{wang2021neus}, \textbf{iButter}~\cite{wang2021ibutter}, \textbf{ST-NeRF}~\cite{zhang2021editable} and \textbf{Neural Body}~\cite{peng2021neural} based on neural radiance field. Note that NeuS only supports static scenes, so we only compare single frame performance with it, the rest of methods support dynamic scenes, we compare the whole sequence with them. For a fair comparison, all the methods share the same training dataset as our approach.
	We choose 90 percent of our captured views as training datasets, and the other 10 percent as novel views for evaluation.
	As shown in Fig.~\ref{fig:compare}, our approach achieves photo-realistic free-viewpoint rendering with the most vivid rendering results in terms of photo-realism and sharpness, which, in addition, can be done in real-time. 
	
	For quantitative comparison, we adopt the peak signal-to-noise ratio (\textbf{PSNR}), structural similarity index (\textbf{SSIM}), mean absolute error (\textbf{MAE}), and Learned Perceptual Image Patch Similarity (\textbf{LPIPS}) \cite{zhang2018perceptual} as metrics to evaluate our rendering accuracy. 

	As shown in Tab.~\ref{table:quantitative comparison_perform}, our approach outperforms other methods in terms of all the metrics for appearance.
	Such a qualitative comparison illustrates the effectiveness of our approach to encode the spatial and temporal information from our multi-view setting.
	
	\begin{figure}[t]
		\begin{center}
			\includegraphics[width=0.98\linewidth]{ 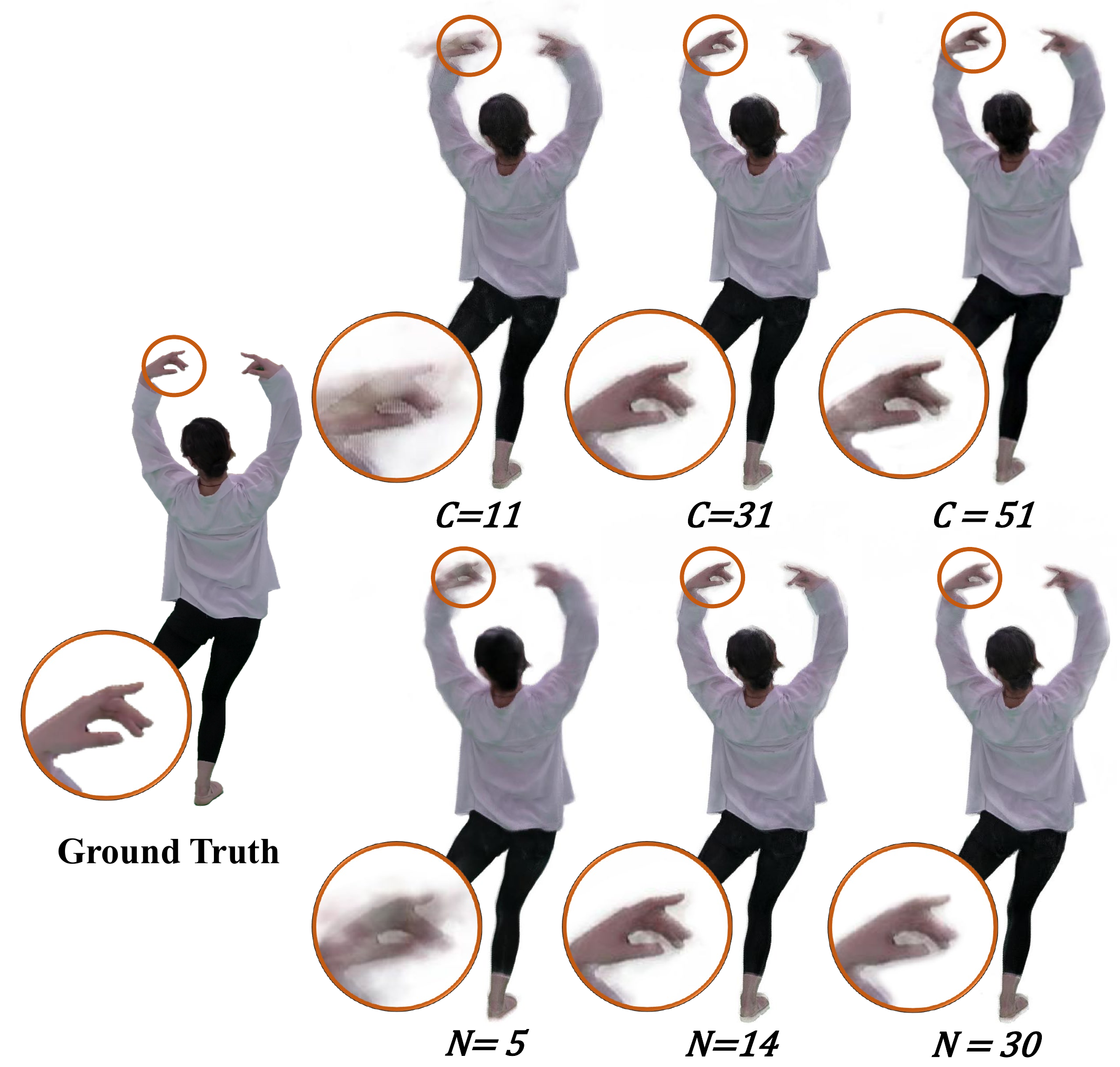}
		\end{center}
		
		\caption{\textbf{Qualitative evaluation on the number of bases and HH dimensions.} The setting with $C=31$ and $N=14$ achieves the satisfactory rendering quality while higher number of bases and HH dimensions does not result in a significant improvement.}
		\label{fig:eval_hh}
		
	\end{figure}
	
	\paragraph{Ablation Study.}
	We first evaluate our two main components in our method, including HH dimensions in hyperspherical harmonic basis function and the number of learnable bases of density and hyper angle. 
	We perform various experiments for different HH dimensions and latent space dimensions and decide the appro choice of the hyperparameters in our algorithm based on the image quality metrics, including \textbf{PSNR}, \textbf{SSIM} and \textbf{MAE} and the \textbf{memory usage} overhead.

	\paragraph{Hyperspherical Harmonic Basis Function.} We first conduct an experiment to search for a compromising HH dimension $N$ in hyperspherical harmonic basis function to balance the realistic rendering performance and memory usage. 

	As shown in Fig. \ref{fig:eval_hh} and Tab. \ref{table:eval_hh}, the results with $HH = 11$ have a better appearance than those using smaller hyperspherical harmonic dimensions and have similar rendering quality and less storage cost than using even higher dimensions. Therefore, $HH = 11$ is a balanced choice on the hyperspherical harmonic basis function.
	
	\paragraph{Number of learnable bases.} We also carry out another experiment to explore the reasonable number of bases $C$ for the time-varying density and hyper angles in Sec.~\ref{sec:npv}.

	As shown in Fig. \ref{fig:eval_hh} and Tab. \ref{table:eval_latent}, the results with the number of bases $C = 31$ have a large improvement compared smaller number of bases, and then continue to increase the bases number has no significant effect on the appearance improvement but increases the memory usage. 
	Our model keeps an outstanding balance.

	\begin{table}[t]
		\caption{\textbf{Quantitative evaluation on the number of learnable base.} Compared with other choices, the setting with $C=31$ achieves the best balance among rendering accuracy, time and storage.}
		\centering
		\resizebox{\linewidth}{!}{ % YANSHUN: scale tabular to fit linewidth
			\begin{tabular}{l|c|c|c|c}
				\multicolumn{5}{c}{ \colorbox{best1}{best} \colorbox{best2}{second-best} } \\
				Latent dimensions     & PSNR$\uparrow$         & SSIM $\uparrow$    & MAE $\downarrow$       & Storage (GB)$\downarrow$ \\ \hline
				$C=11$        & 28.99    & 0.9802 & 0.0067 &\cellcolor{best1}0.716\\
				$C=31$ (ours) & \cellcolor{best2} 31.01  & \cellcolor{best1}0.9856 & \cellcolor{best1}0.0051 & \cellcolor{best2}1.427  \\
				$C=51$       & \cellcolor{best1}31.04 & \cellcolor{best2}0.9854 & \cellcolor{best2}0.0052  & 1.534\\  \hline
		\end{tabular}}
		\rule{0pt}{0.05pt}
		\label{table:eval_latent}
		%\vspace{-3mm}
	\end{table}
	
	\begin{table}[t]
		\caption{\textbf{Quantitative evaluation on Hyperspherical Harmonic Basis Function.} Compared with other choices, the setting with $N = 14$ achieves the best balance among rendering accuracy, time and storage.}
		\centering
		\resizebox{\linewidth}{!}{ 
			\begin{tabular}{l|c|c|c|c}
				\multicolumn{4}{c}{ \colorbox{best1}{best} \colorbox{best2}{second-best} } \\
				Basis     & PSNR$\uparrow$         & SSIM $\uparrow$    & MAE $\downarrow$       & Storage (GB)$\downarrow$ \\ \hline
				$N=5$        & 28.89                  & 0.9823 & 0.0066 & \cellcolor{best1}0.957\\
				$N=14$ (ours) &\cellcolor{best2} 31.01  & \cellcolor{best2}0.9856 & \cellcolor{best2}0.0051 & \cellcolor{best2}1.427  \\
				$N=30$ & \cellcolor{best1} 31.60 &\cellcolor{best1}0.9867 &\cellcolor{best1}0.0048                          & 2.131 \\ \hline
			\end{tabular}
		}
		\rule{0pt}{0.05pt}
		\label{table:eval_hh}
		%\vspace{-3mm}
	\end{table}
	
	\subsection{Composition, Editing, and Lighting Effects}
	
\begin{figure}[t]
		\includegraphics[width=\linewidth]{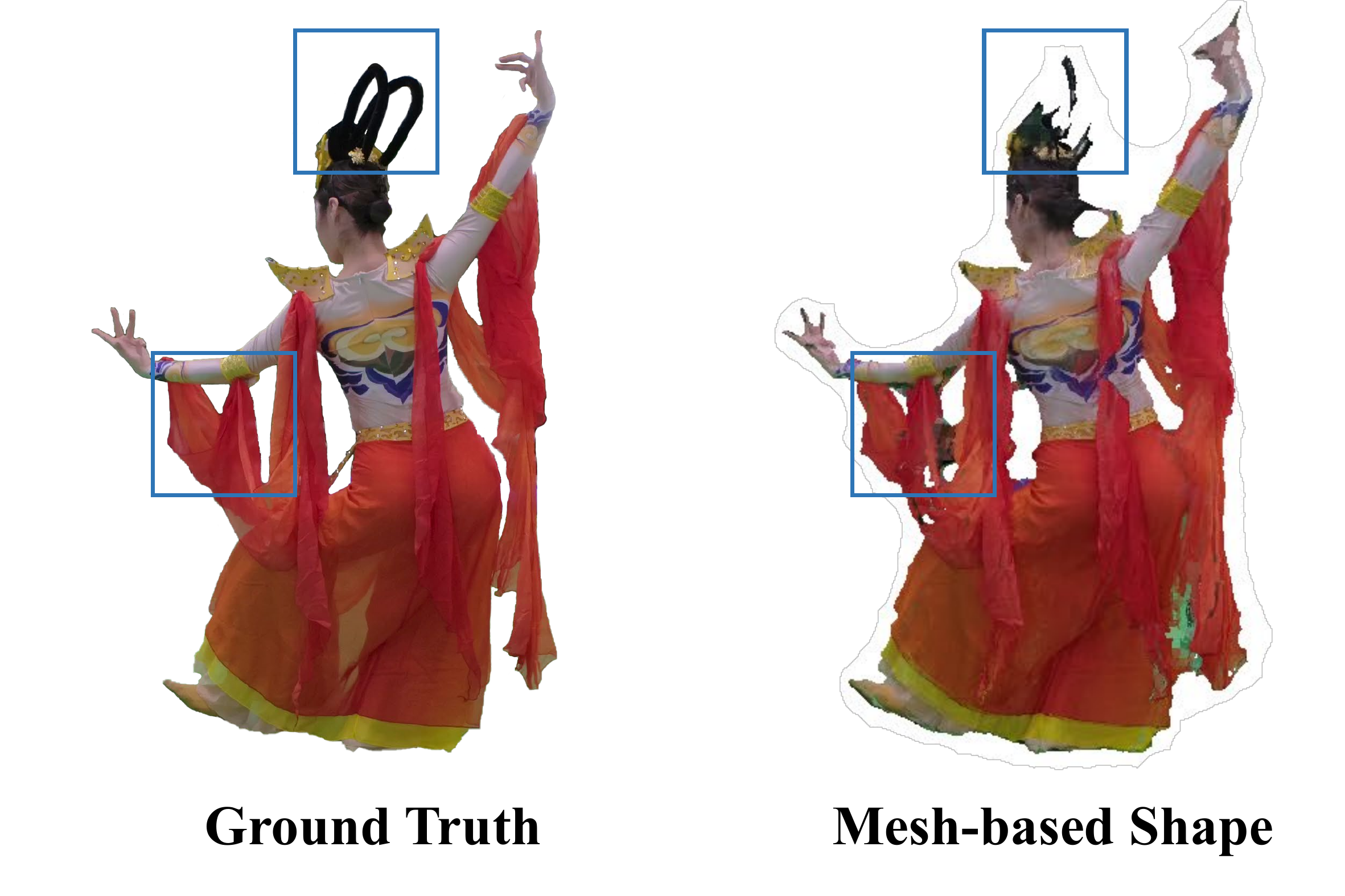}
		\caption{\textbf{Reconstruction Result.} We reconstruct one frame of our captured dataset by RealityCapture\cite{realitycapture}, it cannot handle small, deformable, and semi-transparent geometry.}
		\label{fig:mesh_shape} 
	\end{figure}

	\paragraph{NeuVV vs. 3D Mesh.} Compared with 3D reconstruction methods, NeuVV as a hybrid implicit-explicit representation is particularly useful to handle small, deformable, and semi-transparent geometry. In Dunhuang flying apsaras sequence (Fig.~\ref{fig:mesh_shape}), the performer wears the traditional dancing dress with many long, narrow, thin, and soft ribbons that exhibit complex mutual occlusions. Their geometry and movements are difficult to recover or even manually model using 3D representations. For example, active or passive scanning produces various visual artifacts such as adhesiveness, holes, and noises whereas NeuVV presents a unique advantage by faithfully reproducing plausible rendering at any viewpoint without explicitly revealing the underlying geometry.

	\paragraph{Duplication.} Fig. ~\ref{fig:manifestations} demonstrates how to realize duplicated manifestations of the same Dunhuang dancer. The supplementary video demonstrates how a user creates such effects in virtual space: they first select the VOctree primitive using the controller, then duplicate her multiple times and position individual duplicates at different locations. Finally, they adjust the timing of the movement of each duplicate and hit the play button on the controller to synthesize visual effects similar to the Matrix which used to require professional production. More excitingly, for the first time, a user can view this effect in virtual environments. For example, by positioning the duplications along a line, the front view produces an astounding visual effect of a \textit{Thousand Armed Avalokiteshvara} for conveying the goddess' greatest compassion whereas a side reveals the movements from different perspectives, we show the similar effects in Fig.~\ref{fig:gallery} \textit{Waving} which to create waving effect of the same performer. As aforementioned, duplications do not incur additional memory cost as they share the same VOctree data and therefore it is indeed possible to produce a multiple duplications and still render at an interactive speed.

	\paragraph{Composition.} Composition is a powerful tool in 2D videography. Composition of 3D videos immersive environment is even more exciting. For example, to produce an immersive musical or concert, it is essential to position pre-recorded volumetric performances from different places around the world to the same virtual space. Immersive viewing is achieved via our NeuVV + OpenVR framework that support simultaneously rendering multiple VOctrees of different performers at the same time. The current limit is on GPU memory: each VOctree is about 1-2 GBs and on Nvidia RTX 3090 we can support at most ~12 entities.  Fig.~\ref{fig:teaser} shows an example that we put a ballet performance, a Dunhuang flying apsaras, and modern dance on the same floor. Our spatial-temporal adjustment tools can efficiently synchronize their movements where our depth sorted rendering manages to produce correct occlusions as the viewer changes position in virtual space. Since VOctree presents a neural volume representation with opacity, translucency can achieve partial see-through effects. 
	
	\paragraph{Free-viewpoint Video.} A byproduct of our real-time, multi-VOctree rendering is the acceleration of free-viewpoint video (FVV) production. Existing FVVs, especially the neural ones~\cite{zhang2021editable}, are produced offline. By providing real-time rendering capability, NeuVV enables live feedback to the videographer, who can adjust the position, size, and timing of the contents on the fly, greatly improving production efficiency. With the support the latest near real-time neural technique such as NGP~\cite{muller2022instant}, live performance composition and editing in the form of NeuVVs may be practical in foreseeable future. As illustrated in Fig.~\ref{fig:gallery} \textit{When Van Gogh Meets Ballet}, we show representative frames in rendered FVV using edited VOtree, the edited results achieve more artistic effects.
	
	\paragraph{Lighting.} Traditionally lighting effects are achieved on explicit geometry such as meshes. As a hybrid neural-volume representation, VOctree-based shadowing and falloff estimation (Section \ref{sec:immersive-editing}) can produce certain lighting effects. Fig.~\ref{fig:gallery} shows the lighting effect by positioning a point light source on side of performer were the cast shadow serves as virtual motion magnifier. Nuances in small movements such as hands and arms as well as clothes deformation are better illustrated through time-varying shadows in real-time, adding another layer of realism as if in real theaters. Falloff lighting further helps guide the viewer's focus on different parts and produce smooth transitions to real or synthetic background. Both shadows and fallout lighting can be conducted in one pass via the estimation of the alpha/depth map of VOctree, and by using a low resolution shadow/depth map, they reduce the overall rendering speed (from the viewer's perspective)  by about 20\%. More advanced shading that requires using surface normal, however, is not readily available in the current representation, although latest extensions such as NeuS\cite{wang2021neus} may be integrated into VOctree as a potential remedy. 
	
	\paragraph{Interaction.} As the final example, we combine the motion of the viewer with the performer in the experience of VR fitness training. One of the most exciting experiences Metaverse promises is to offer live interactions with virtual characters in virtual environments. In this specific case, a user should not only be able to omnidirectionally watch the virtual trainer's moves but also compare their own moves with the trainer. In our implementation, we use a single camera motion capture solution~\cite{he2021challencap} that estimates 3D skeleton structures of users as they move. We also precompute the "ground truth" skeleton moves of the trainer, by first rendering a multi-view video of whole body movements also using NeuVV and then conducting multi-view skeleton estimation. Finally we highlight skeleton discrepancies between the two on top of NeuVV rendering, to remind the user about incorrect postures. The user can then pause and move about the trainer with the right perspective for a replay. 
	
	\subsection{Possible Extensions}
	NeuVV is designed to produce high quality multi-view video rendering instead of 3D reconstruction, and therefore it cannot yet produce satisfactory geometry from VOctree. Brute-force approaches such as converting per-frame density field to meshes via thresholding and marching cubes lead to pure reconstruction, especially under fast motions. This should be viewed as a limitation as the results cannot be readily integrated into existing production and rendering pipelines such as Unity, Unreal, Blender, etc., that still rely on mesh inputs. Because the support for neural rendering is provided on these engines, a possible extension is to resort to traditional or neural geometric modeling tools. 
	
	For example, one can render foreground maps at an ultra dense set of views and use the masks to conduct space carving. Alternatively, recent approaches based on signed distance functions (SDF) such as NeuS~\cite{wang2021neus} may be integrated into the NeuVV pipeline. 
	
	Same as existing neural approaches for handling dynamic objects, we use relatively short footage (around 3$\sim$6 seconds). The challenges are multi-fold. Longer clips correspond to longer training time and higher storage. In particular, as NeuVV optimizes over all frames from all viewpoints, the memory limit on the GPU restricts the length of the footage. Speed and memory aside, long sequences may produce very large motions that cannot be fully capture by HH and our learnable scheme. 
	
	One potential solution is to borrow the idea of keyframe based video compression where the video can be truncated into smaller pieces, each individually compressed or trained in our case. In video compression, only changes that occur from one frame to the next are stored in the data stream. It is possible that we can apply NeuVV training only on the residues, e.g., by pre-processing videos at individual viewpoint and set out to optimize the changes rather than the complete frames. Such a scheme may also provide a viable streaming strategy of NeuVV and is our immediate future work.
	
	Though our NeuVV exhibits capacity in photo-realistic rendering and editing of volumetric video content in real-time, there are several limitations and consequently possible extensions to our approach. Firstly, our NeuVV is a NeRF based representation, compared to NeRF's compelling novel view synthesis ability, the geometry recovered is general lower quality. 
	
	Similarly, our NeuVV suffers the same geometry recovery problem given a static time frame. Moreover, the recovered geometries exhibits ghosting effect when the performer's motion is too fast. This is because the change of volume density is constraint by learnable bases, which can well handle smooth motion but reluctant to fast density changes. 
	
	The lack of high quality geometry greatly limits the application of NeuVV as current industrial graphics rendering engines, such as OpenGL and Unity3D, only support a mesh based geometry representation. Before a natural integration of neural rendering into traditional rendering engines, an possible extension is resorting to cooperate with stronger geometry recovery approaches, such as the signed distance function (SDF) and neural graphic primitives.
	
	Moreover, all videos demonstrations in our paper are relatively short (around 3$\sim$6 seconds) as NeuVV is more difficult to converge when the input video is long. Also we may have to sacrifice some storage for high quality rendering as motions in longer videos are likely to be complicated and we have to use higher dimensions of the latent space to account for the complex motion. 
	
	We can borrow the concept of key frames in video compression to potentially solve this problem. Particularly, we can separate a long video into small segments, and each segment is defined by a key frame. Within each segment, motion of the performer is relatively small. And hence we can optimize one NeuVV for each segment effectively.
	
	Finally, transferring NeuVV over internet is not efficient as we have to send the whole volume representation at once, no matter which frame is of the viewer's interest. One possible solution is to directly slice the VOctree at a given time frame to obtain the SH coefficient, and transform the the time frame into a PlenOctree representation and then compress and transmit over internet.

	\section{CONCLUSION}
	
	We present a new neural volumography technique, NeuVV, which leverages the neural rendering technique to tackle volumetric videos. We model the scene captured by a volumetric video as a dynamic radiance field function, which maps a 6D vector (3D position + 2D view direction + 1D time) to color and density. Our NeuVV encodes a dynamic radiance field effectively, as the core at our NeuVV is a factorization schemes by hyperspherical harmonics, to account for the angular and temporal variations at each position. Density at a specific position only exhibits temporal variations while being invariant to view directions. Hence we further develop a learnable basis representation for temporal compaction of densities. Similar to the PlenOctree~\cite{yu2021plenoctrees}, our NeuVV can be easily converted into an octree based representation, which we call VOctree, for real-time rendering and editing. NeuVV tackles a volumetric video sequence as a whole therefore reduces the memory overhead and computational time by two orders of magnitudes.
	
	For demonstration, we further provides tools based on NeuVV for flexibly composing multiple performances in 3D space, enabling interactive editing in both spatial and temporal dimensions, and rendering a new class of volumetric special effects with high photo-realism. More specifically, we demonstrate that NeuVV, or VOctree more precisely, allows for real-time adjustments of the 3D locates, scales of multiple performers, re-timing and thus coordinating the performers, and even duplicating the same performer to produce varied manifestations in space and time. 
	To the best of our knowledge, NeuVV is the first neural based volumography technique that supports real-time rendering and interactive editing of volumetric videos.

	\bibliographystyle{ACM-Reference-Format}
	\bibliography{ref}
	
	\appendix
	\section*{APPENDIX}
	\setcounter{section}{1}
	\subsection{Complex HH in 4D Hyperspherical Coordinates}
	
	Hyperspherical Harmonics are widely used in quantum mechanical and chemistry field to solve few-body systems. It also has been used in computer graphics visualization 
	% 3D shape descriptors: 4D hyperspherical harmonics “An exploration into the fourth dimension”
	~\cite{10.5555/1712936.1712961} and the representation of complicated brain subcortical structures~\cite{PASHAHOSSEINBOR201589}
	
	4D complex Hyperspherical Harmonics can be derived from as complex 3D Shpherical Harmonics~\cite{spherical} 
	\begin{equation}\label{eqn:complex-HSH-app}
		\mathcal{H}^m_{nl}(\theta, \phi, \gamma) = A_{n,l} \sin^l(\gamma) C^{l+1}_{n-l} \big (\cos(\gamma) \big ) \mathcal{S}^m_l(\theta, \phi)
	\end{equation}
	where 
	\begin{equation}
		A_{n,l} = (2l)!! \sqrt{\frac{2(n+1)(n-l+1)!}{\pi (n+l+1)!}}
	\end{equation}
	
	\noindent $\gamma, \theta \in [0,\pi]$, $\phi \in [0, 2\pi]$ , $C^{l+1}_{n-1}$ are Gengenbauer polynomials, and $\mathcal{S}^m_l$ are the 3D spherical harmonics. $l, m, n$ are integers, where $l$ denotes the degree of the HH, $m$ is the order, and $n = 0, 1, 2, ...$, following $0 \leq l \leq n$ and $-l \leq m \leq l$.
	
	3D spherical harmonics $Y^m_l(\theta, \phi)$ are defined as below:
	
	\begin{equation}
		Y^m_l(\theta, \phi) = K^m_l P^m_l(\cos\theta)e^{im\phi}
	\end{equation}
	where 
	\begin{equation}
		K^m_l = (-1)^m\sqrt{\dfrac{2l+1}{4\pi}\dfrac{(l-m)!}{(l+m)!}}
	\end{equation}
	and $P^m_l$ is associated Legendre polynomials.
	
	\subsection{Real-valued HH in 4D Cartesian Coordinates}
	It is hard to directly use HH in complex space in our approach as it has a heavy burden for calculating its imaginary part and optimizing our network weights by traditional grad descent methods. Thus, we derived how to transform a 4D complex HH to be in real space. We have implemented a program to iteratively solve and verify $N-dimensional$ HH basis function. And we will release the code in the future.
	
	real-valued HH in 4D Cartesian space input a 4D unit vector $\mathbf{x} = [x_1, x_2, x_3, x_4]^T$, the relationship between $\mathbf{x}$ and $(\gamma, \theta, \phi)$ is as below:
	
	\begin{eqnarray} \label{eqn:var_relation}
		x_1 &=& \cos(\gamma) \nonumber\\
		x_2 &=& \sin(\gamma)\cos(\theta) \nonumber\\
		x_3 &=& \sin(\gamma)\sin(\theta)\cos(\phi) \nonumber\\
		x_4 &=& \sin(\gamma)\sin(\theta)\sin(\phi) \nonumber\\
	\end{eqnarray}
	
	where $\sum_{i=1}^{n}x_i^2 = 1$.
	
	The real-valued SH $Y_{lm}$ has been given as \cite{BLANCO199719}
	\begin{equation}\label{eqn:real-SH}
		Y_{lm} = \left\{\begin{aligned}
			\frac{1}{\sqrt{2}}\left(Y^m_l + (-1)^m Y^{-m}_l\right) \quad \text{if }m > 0 \\
			Y^m_l \quad \text{if } m = 0 \\
			\frac{1}{\sqrt{2}}\left(Y^{-m}_l - (-1)^m Y^{m}_l\right) \quad \text{if }m < 0 \\
		\end{aligned}
		\right.
	\end{equation}
	
	We observe that the similar idea can be used to obtain real-valued HSH $\mathcal{H}_{nlm}(\theta, \phi, \gamma)$ as the complex number is only from $Y^m_l$, combine Eqn.~\ref{eqn:complex-HSH-app} with Eqn.~\ref{eqn:real-SH}:
	\begin{equation}
		\mathcal{H}_{nlm}(\theta, \phi, \gamma) = 
		A_{n,l}sin^l(\gamma)C^{l+1}_{n-l}(\cos(\gamma))Y_{lm}(\theta, \phi)
	\end{equation}
	
	Finally, to transform 4D hypersphere coordinates to 4D Cartesian coordinates,
	we then substitute $(\gamma, \theta, \phi)$ with $\mathbf{x}$. Using the same definition of Eqn.~\ref{eqn:var_relation}. We further introduce a separated Cartesian form of $Y_{lm}(x_2, x_3, x_4)$ in 3D Cartesian coordinates.
	
	\begin{equation}
		\begin{bmatrix}
			Y_{lm} \\
			Y_{l-m}
		\end{bmatrix} = \sqrt{\dfrac{2l+1}{4\pi}}\Bar{\prod}^m_l(x_2)
		\begin{bmatrix}
			A_m \\
			B_m
		\end{bmatrix}, m > 0
	\end{equation}
	\begin{equation}
		Y_{l0} = \sqrt{\dfrac{2l+1}{4\pi}}\Bar{\prod}^m_0(x_2)
	\end{equation}
	where
	\begin{eqnarray}
		A_m(x_3, x_4) &=& \sum^m_{p=0}\tbinom{m}{p}x_3^p x_4^{m-p} \cos((m-p)\frac{\pi}{2}) \\
		B_m(x_3, x_4) &=& \sum^m_{p=0}\tbinom{m}{p}x_3^p x_4^{m-p} \sin((m-p)\frac{\pi}{2})
	\end{eqnarray}
	and
	\begin{eqnarray}
		\Bar{\prod}^m_l(x_2)& = &\sqrt{\dfrac{(l-m)!}{(l+m)!}}
		\sum\limits^{\lfloor (l-m)/2\rfloor}_{k=0} B_{k,lm} x_2^{l-2k-m}\nonumber\\
		B_{k,lm} &=& (-1)^k2^{-l}\tbinom{l}{k}\tbinom{2l-2k}{l}\dfrac{(l-2k)!}{(l-2k-m)!} \\
	\end{eqnarray}
	
	Finally, we have:
	\begin{equation}\label{eqn:real-valued-HSH-1}
		\begin{bmatrix}
			\mathcal{H}_{nlm} \\
			\mathcal{H}_{nl-m}
		\end{bmatrix} = A_{n,l}(1-x_1^2)^{l/2}C^{n+l}_{n-l}(x_1)\begin{bmatrix}
			Y_{lm} \\
			Y_{l-m}
		\end{bmatrix}, m > 0
	\end{equation}
	When $m=0$,
	\begin{equation}\label{eqn:real-valued-HSH-2}
		\mathcal{H}_{nl0} = A_{n,l}(1-x_1^2)^{l/2}C^{n+l}_{n-l}(x_1)\cdot\sqrt{\dfrac{2l+1}{4\pi}}\Bar{\prod}^0_l(x_2)
	\end{equation}
	Using Eqn.~\ref{eqn:real-valued-HSH-1} and Eqn.~\ref{eqn:real-valued-HSH-2}, we can derive the simplest forms of HH basis. The similar idea can be used to derive more higher dimensional HH basis functions.

	% \begin{figure}
	% \centering
	% \includegraphics[width=1.0\textwidth]{table.png}
	% \caption{\label{fig:HSH_table} Table of HH coefficients. }
	% \end{figure}

\end{document}